\documentclass[acmsmall]{acmart}

\usepackage{algpseudocode}
\usepackage{algorithm}
\usepackage{graphicx}
\usepackage{subcaption}
\usepackage{subfig}
\usepackage{color}
\usepackage{soul}
\usepackage{amsmath}

\usepackage{amssymb}
\usepackage{multirow}
\usepackage[final, commandnameprefix=ifneeded]{changes}
\usepackage{pifont}
\newcommand{\greenup}{\textcolor{green}{\ding{115}}} % upward green triangle
\newcommand{\reddown}{\textcolor{red}{\ding{116}}}   % downward red triangle
\newcommand{\greendown}{\textcolor{green}{\ding{116}}} % downward green triangle

\AtBeginDocument{%
  }

\setcopyright{acmlicensed}
\acmJournal{PACMHCI}
\acmYear{2025} \acmVolume{9} \acmNumber{2} \acmArticle{CSCW084} \acmMonth{4}\acmDOI{10.1145/3710982}

\makeatletter
\providecommand{\sf@counterlist}{}
\makeatother

\begin{document}

%%
%% The "title" command has an optional parameter,
%% allowing the author to define a "short title" to be used in page headers.
\title[FairPlay]{\textsc{FairPlay}: A Collaborative Approach to Mitigate Bias in Datasets for Improved AI Fairness}

%%
%% The "author" command and its associated commands are used to define
%% the authors and their affiliations.
%% Of note is the shared affiliation of the first two authors, and the
%% "authornote" and "authornotemark" commands
%% used to denote shared contribution to the research.
\author{Tina Behzad}
\email{tbehzad@cs.stonybrook.edu}
\orcid{0009-0009-1157-9082}
\authornote{Authors contributed equally to this research.}
\affiliation{%
  \institution{Department of Computer Science, Stony Brook University}
  \city{Stony Brook}
  \state{New York}
  \country{USA}
}

\author{Mithilesh Kumar Singh}
\email{mkssingh@cs.stonybrook.edu}
\orcid{0009-0007-6477-1495}
\authornotemark[1]
\affiliation{%
  \institution{Department of Computer Science, Stony Brook University}
  \city{Stony Brook}
  \state{New York}
  \country{USA}
}

\author{Anthony J. Ripa}
\email{aripa@cs.stonybrook.edu}
\orcid{0000-0002-1412-4384}
\authornotemark[1]
\affiliation{%
  \institution{Department of Computer Science, Stony Brook University}
  \city{Stony Brook}
  \state{New York}
  \country{USA}
}

\author{Klaus Mueller}
\email{mueller@cs.stonybrook.edu}
\orcid{0000-0002-0996-8590}
\affiliation{%
  \institution{Department of Computer Science, Stony Brook University}
  \city{Stony Brook}
  \state{New York}
  \country{USA}
}
%%%%
% \author{Mithilesh Kumar Singh}
% \email{mkssingh@cs.stonybrook.edu}
% \orcid{0009-0007-6477-1495}

% \author{Anthony J. Ripa}
% \email{aripa@cs.stonybrook.edu}
% \orcid{0000-0002-1412-4384}

% \author{Tina Behzad}
% \email{tbehzad@cs.stonybrook.edu}
% \orcid{0009-0009-1157-9082}

%%
%% By default, the full list of authors will be used in the page
%% headers. Often, this list is too long, and will overlap
%% other information printed in the page headers. This command allows
%% the author to define a more concise list
%% of authors' names for this purpose.
% \renewcommand{\shortauthors}{Trovato et al.}
\renewcommand{\shortauthors}{Behzad, Singh, Ripa, and Mueller}

%%
%% The abstract is a short summary of the work to be presented in the
%% article.
\begin{abstract}
  The issue of fairness in decision-making is a critical one, especially given the variety of stakeholder demands for differing and mutually incompatible versions of fairness. Adopting a strategic interaction of perspectives provides an alternative to enforcing a singular standard of fairness. We present a web-based software application, FairPlay, that enables multiple stakeholders to debias datasets collaboratively. With FairPlay, users can negotiate and arrive at a mutually acceptable outcome without a universally agreed-upon theory of fairness. In the absence of such a tool, reaching a consensus would be highly challenging due to the lack of a systematic negotiation process and the inability to modify and observe changes. We have conducted user studies that demonstrate the success of FairPlay, as users could reach a consensus within about five rounds of gameplay, illustrating the application's potential for enhancing fairness in AI systems.
\end{abstract}

%%
%% The code below is generated by the tool at http://dl.acm.org/ccs.cfm.
%% Please copy and paste the code instead of the example below.
%%


%%
%% Keywords. The author(s) should pick words that accurately describe
%% the work being presented. Separate the keywords with commas.

\received{January 2024}
\received[revised]{July 2024}
\received[accepted]{October 2024}

%%
%% This command processes the author and affiliation and title
%% information and builds the first part of the formatted document.
\maketitle

\section{Introduction}
Fairness remains an elusive goal in our increasingly data-driven world, hindered by the Impossibility of Fairness \cite{Friedler21}. This paradox emerges from the diversity of ideological beliefs surrounding the concept of fairness, creating a scenario where achieving a universally agreed-upon definition becomes unfeasible. \par

Although the literature has defined a myriad of notions to quantify fairness, each measures and emphasizes different aspects of what can be considered “fair”. Many are difficult/impossible to combine \cite{10.1257/pandp.20181018}\cite{doi:10.1089/big.2016.0047}, but ultimately, we must keep in mind (as noted in \cite{pmlr-v89-chierichetti19a}) there is no universal means to measure fairness, and at present no clear guideline(s) on which measures are “best” \cite{10.1145/3616865}. \par
This problem's essence is deeply rooted in context-specific nuances, making it crucial to tailor solutions to the individual characteristics and challenges of each case. Consequently, it becomes vital for human experts to define what constitutes fairness in each distinct scenario. As the range of situations where models are deployed for decision-making expands, so does the necessity for a diverse group of people to scrutinize these models for fairness. To facilitate this, a variety of interfaces have been created, enabling experts across disciplines to assess different fairness metrics and determine the best strategies for mitigating bias in datasets or models \cite{10.1145/3411764.3445604}. These tools are designed to empower those with in-depth knowledge in their respective fields to define and implement fairness in their models. However, a notable gap in these tools is the lack of a collaborative approach in the bias mitigation activities. \par
Our approach is rooted in a more practical and collaborative method, inspired by the practice of negotiation for consensus building. We acknowledge and utilize ideological diversity as a strength, channeling it to bring together various stakeholders to collectively define fairness for their specific task. We build our software on the foundation of a previously published web-based software, D-BIAS \cite{ghai2022d}. D-BIAS is a visual and interactive human-in-the-loop method designed for the pre-processing phase of debiasing algorithmic decision systems (ADS) by ways of a causal model initially derived from the original (potentially biased) training data. Unlike traditional metric-driven methods, D-BIAS provides a detailed view of variable interactions and their impact on the outcome, enabling users to make modifications. 

Our system features an enhanced web interface that shifts D-BIAS from a single-user mode to a multi-user framework. Here, professionals from different fields or different stakeholders work together to identify the most fitting and fair causal structure for their specific domain scenario, promoting a consensus-based methodology. Then, once consensus has been reached, the debiased data generated by the causal model can be used to train any ADS. This evolution leads to a more dynamic and inclusive strategy for achieving fairness, where all stakeholders or experts can collaboratively arrive at a mutually satisfactory resolution. 

To evaluate our system, we conducted user studies with four different groups, analyzing the effectiveness of our method in reaching consensus on a hiring dataset. The studies reveal that users typically agree to end the game after about five rounds of gameplay, indicating collective agreement on the final causal model.

Our research contributions are:

\begin{itemize}
  \item The first collaborative methodology and tool for the debiasing of ADS training data, to the
    best of our knowledge.
  \item Evolving D-BIAS into a multi-user tool for domain stakeholders to reach consensus. 
  \item  Expanding the D-BIAS interactive visual interface by a set of new visual feedback widgets, specifically designed to help stakeholders track their progress more effectively.
  \item Executing a user study with four groups, each comprising five stakeholders, centered around a hiring ADS scenario.
  \item Analyzing observed user behavior patterns from the study.
  \item Evaluating the gathered debiasing outcomes using standard metrics.
  \end{itemize}
In this paper, we present our structured collaborative method in the form of a game, and throughout the text, we use the terms 'users', 'players', and 'stakeholders' interchangeably.

\section{Related Work}

In this section, we review existing approaches to fairness in machine learning, discuss interactive approaches for addressing bias and fairness, explore consensus-building mechanisms in related domains, and highlight the importance of visualizing these techniques to enable broader understanding and adoption.\par
%Fairness in machine learning has been extensively studied, and various approaches have been proposed to address bias and discrimination in algorithmic decision-making.\par
\subsection{Fairness in Machine Learning}

\paragraph{\added{Challenges in Defining Fairness Metrics}} Different fairness metrics and definitions have been developed to quantify and measure bias in machine learning models \cite{10.1145/3616865}\cite{10.1145/3494672}. Commonly used fairness metrics include demographic or statistical parity \cite{dwork2012fairness}\cite{Kleinberg16}\cite{binns2018fairness}, equal opportunity \cite{hardt2016equality} \cite{agarwal2018reductions}, and equalized odds \cite{dwork2012fairness} \cite{hardt2016equality}. These metrics provide quantitative measures to assess the fairness of decisions made by the models across different groups. Different notions and measures can be mutually incompatible and entail unavoidable tradeoffs~\cite{kleinberg2016inherent}\cite{Friedler16}.  There is no consensus on a single most appropriate definition of fairness  \cite{gajane2017formalizing}. Determining the right measure to be used must take into account the proper legal, ethical, and social context \cite{10.1145/3494672}. For a given application in a given context, algorithms can not be expected to determine the most appropriate definition of fairness and decide a desirable tradeoff between different metrics that is acceptable to all stakeholders. On the other hand, a human trusted by the majority of stakeholders can make an informed decision when presented with the required information \cite{shneiderman2022human}. Hence, introducing a human in the loop can improve perceived fairness. As for the aspect of trust, people are more likely to trust a system if they can tinker with it, even if this means making it perform imperfectly 
 \cite{dietvorst2018overcoming}.\par
 
\paragraph{\added{Bias Mitigation Visualization Tools}} Understanding and interpreting these fairness approaches might be challenging, especially for non-experts or individuals without a strong technical background such as the stakeholders in a given task. Therefore, in recent years, efforts to visualize and explain these techniques have been developed \cite{10.1145/3411764.3445604}. Some of these methods include: Silva \cite{yan2020silva}, FairVis \cite{cabrera2019fairvis}, FairRankVis \cite{xie2021fairrankvis}, DiscriLens~\cite{wang2020visual}, FairSight~\cite{ahn2019fairsight}, What-If toolkit (WIT) \cite{wexler2019if}, Aequitas\cite{saleiro2018aequitas}, AI Fairness 360 (AIF360) \cite{Fairness360} and D-BIAS \cite{ghai2022d}. These tools are crucial to enable a broader audience to understand and engage with fairness in machine learning.  Most of these tools focus on bias identification. Some of them, such as FairSight and AIF360, also permit debiasing. D-BIAS, which this paper is built upon, is similar to Silva which also features a graphical causal model in its interface. Silva’s empirical study showed that users can interpret causal networks and found them helpful in identifying algorithmic bias \cite{yan2020silva}. However, like most other visual tools, Silva is limited to bias identification. D-BIAS presents a tool that supports both bias identification and mitigation using a graphical causal model. \par

\paragraph{\added{Approaches to Achieving Fairness}} Various approaches have been proposed to achieve fairness in machine learning.\cite{10.1145/3616865}\cite{10.1145/3494672} Pre-processing methods focus on modifying the training data to remove bias before training the model \cite{calmon2017optimized}\cite{kamiran2012data}. In-processing methods aim to modify the learning algorithm or objective function to directly optimize for fairness \cite{wan2023processing}. Post-processing methods modify the model's predictions after training to achieve fairness \cite{kim2019multiaccuracy}. Research has shown that teams typically look to their training datasets, not
their ML models, as the most important place to intervene
to improve fairness in their products \cite{10.1145/3290605.3300830}\cite{10.1145/3411764.3445604}. D-BIAS and hence our work relates closely with the pre-processing stage where we make changes to the output label based on users’ decisions.\par

\subsection{Consensus-Building Mechanisms} \label{sec:relatedwork-CRP}

\paragraph{\added{Defining Consensus}} In the literature, two types of consenses are defined \cite{ZHANG2019580}. Several researchers define consensus as the full and unanimous agreement of all the decision-makers regarding all the feasible alternatives \cite{BEZDEK1978255}. However, unanimity may be difficult to achieve, in particular with large and diversified groups of decision-makers as is the case in real-world settings. In contrast, the concept of consensus has also been considered in a more flexible way with regard to its measurement, which has led to the proposal and use of “soft” consensus degrees \cite{CHICLANA2013110} with the aim of achieving two important goals: (i) to reflect better partial agreement; and (ii) to guide the Consensus Reaching Process (CRP) until an acceptable high level of agreement is achieved among decision-makers \cite{ZHANG2019580}.

\paragraph{\added{Challenges in Reaching Consensus}} Consensus-building mechanisms have been extensively studied in fields such as multi-agent systems \cite{li2019survey}, social choice theory \cite{lu2011budgeted},  and deliberative decision-making \cite{van2006making} to address challenges. The most important challenge is the aim to reach an agreement or consensus among multiple stakeholders with diverse preferences and perspectives. Deliberative decision-making frameworks involve structured dialogue and information sharing among stakeholders to collectively arrive at decisions. Innes argues a number of conditions need to hold for a process to be labeled consensus building \cite{innes2004consensus}. If these do not hold, failure of various kinds is likely. Among these conditions are: including a full range of stakeholders, meaningfulness of the task to all participants, mutual understanding of interests, a dialogue where all are heard and respected, a self-organizing process, and accessible information.
To aid in reaching the aforementioned conditions, visualizations have been extensively used to provide a graphical representation of the deliberative process, illustrating arguments, preferences, and their evolution over time \cite{zilouchian2011ideatracker}\cite{feick1999consensus}\cite{eppler2004facilitating}.\par

\paragraph{\added{Incorporating Diverse Perspectives on Fairness}} The lack of tools focusing on a collaborative approach in fairness visualization is a significant gap, particularly considering the critical role collaboration plays in this field \cite{collaborative-design}. Fairness is a concept that varies greatly depending on the context and individual perspectives \cite{context-specific}; a single person or a non-interactive tool might overlook these nuances. Therefore, the use of collaborative visualization tools is essential, as they can amalgamate a range of viewpoints. This integration leads to a more holistic comprehension and implementation of fairness, tailored to the specific requirements of any given task. \par

\paragraph{\added{Gamification and Cooperative Strategies}} Integrating cooperative elements and gamification, as demonstrated in the FairPlay game discussed in this paper, drives active participation and collective decision-making \cite{collaborative-design}. These strategies do more than just engage individuals; they create an environment where the pursuit of fairness becomes a shared goal and players recognize that others understand their positions. According to Scheff \cite{consensus-scheff}, these two elements are essential for reaching consensus. In the absence of tools that focus on collaborative approaches, there is a missed opportunity for richer, more inclusive discussions and solutions around bias mitigation. Such tools could promote a deeper understanding and more effective implementation of fairness in machine learning systems by leveraging the collective intelligence and insights of a wider group of stakeholders \cite{binns2018fairness}.

%%%%%%%%%%%%%%%%%%%%%%%%%%%%%%%
\section{FairPlay Game Design}
\label{GameDesign}

The transformation of the D-BIAS platform into FairPlay, a collaborative environment, involved a design process aimed at creating an engaging experience. This redesign aligns with the guidelines set forth in the fairness toolkit rubric \cite{10.1145/3411764.3445604} and incorporates elements identified as crucial for successful consensus-building in related work \cite{innes2004consensus}. The game mechanics and user interface were thoughtfully developed to emphasize interactivity and gamification, creating a cooperative environment where players can actively engage in modifying the causal graph. \added{A key focus was on developing an intuitive interface that balances complexity with usability, ensuring that even non-technical users can participate. This approach was chosen to make the system accessible to diverse groups, empowering all players to make informed decisions, regardless of their technical expertise.} The structure of the game encourages a self-organizing approach, ensuring that each stakeholder is heard and their metrics and evaluations are accessible to all, promoting mutual understanding of interests.

% tag: Q1
In a real-world scenario, FairPlay can be used by up to five key stakeholders involved in any decision-making process. We chose this number as it provided a good compromise between diversity and manageability; a smaller or larger number of players is also possible. The idea is that each stakeholder brings unique perspectives and concerns, making participation crucial for achieving a fair and unbiased decision. For example in urban planning, five stakeholders could include representatives from city planners, community members, business owners, environmental groups and transportation authorities. In healthcare policy development, stakeholders could be healthcare providers (doctors, nurses, etc.), healthcare administrators, patients, insurance companies, and government regulators.

We demonstrate FairPlay using a hiring task for a programming position, with stakeholders including the hiring agency, the employer, the manager, coworkers, and the union representative. The hiring agency is responsible for matching qualified candidates with job vacancies across various companies. The employer sets the overall hiring policies and goals for the organization. The manager is directly involved in evaluating and selecting candidates, considering factors such as team dynamics and job requirements. Coworkers provide insights into the day-to-day impact of hiring decisions on the work environment and company culture. Finally, the union representative advocates for fair hiring practices and protects the interests of workers. By bringing these five stakeholders together in a collaborative debiasing process, FairPlay ensures that the resulting dataset reflects a balanced and inclusive approach to hiring, taking into account the diverse viewpoints and priorities of each group. It is important to note that while different stakeholders may hold varying levels of influence or power in different scenarios, FairPlay treats all stakeholders as equals, valuing each participant's preferences equally.

The configuration page of the game adheres to the rubric's recommendation of being "Applicable to a diverse range of predictive tasks". The inclusion of visualizations, like causal network diagrams and various charts, aligns with the rubric's focus on providing both a comprehensive and nuanced perspective on fairness and supporting intersectional analysis. This section goes into the specific design features of FairPlay, highlighting its differences from D-BIAS and the rationale behind these changes. A summary of these differences is available in Table \ref{tab:comparison}.

\begin{table}[h]
\resizebox{0.6\textwidth}{!}{
\centering
\scriptsize
\begin{tabular}{|c|c|c|}
\hline
\textbf{Features} & \textbf{D-Bias} & \textbf{FairPlay} \\ \hline
Multi-player & x & \checkmark \\ \hline
Edge modification & \checkmark & \checkmark \\ \hline
Adding/Deleting/Reversing edges & \checkmark & x \\ \hline
Utility and Bias metrics plots & \checkmark & x \\ \hline
Modification History & x & \checkmark \\ \hline
Game metrics, scores and charts & x & \checkmark \\ \hline
% \textbf{Finding paths} & \checkmark & x \\ \hline
\end{tabular}}
\caption{Comparison of Features between D-Bias and FairPlay}
\label{tab:comparison}
\end{table}

\begin{figure}[h]
  \centering
  \includegraphics[width=0.6\linewidth]{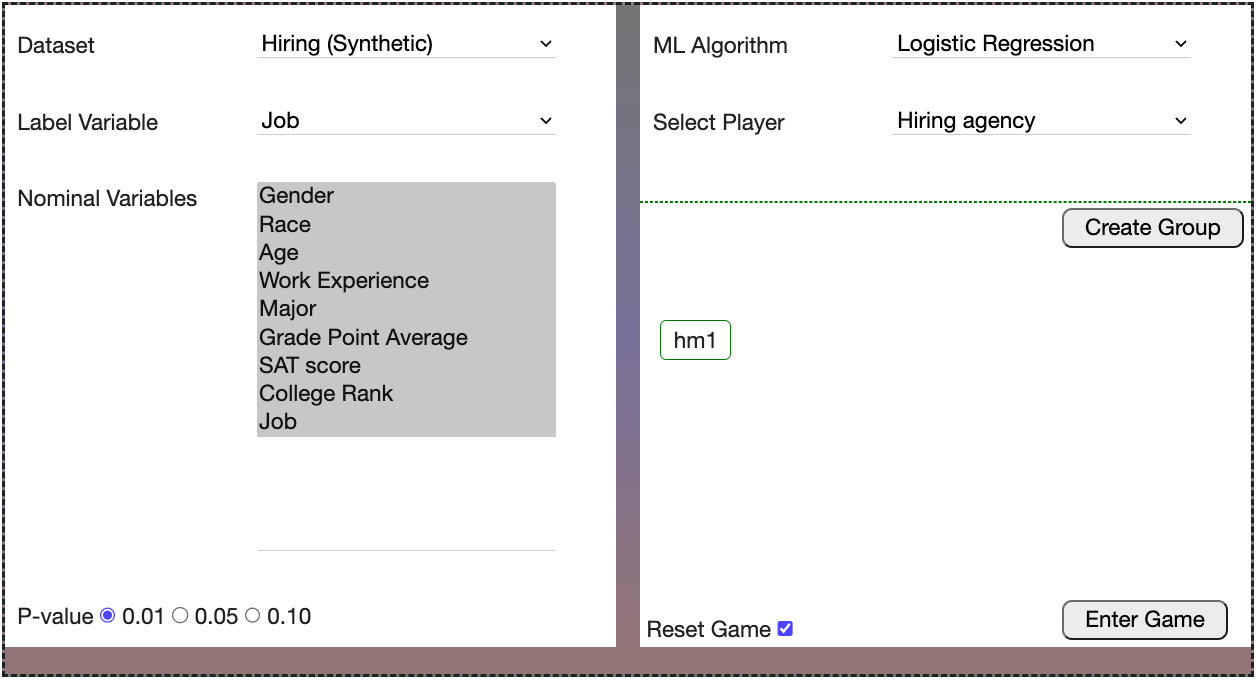}
  \caption{FairPlay: Game Configuration. This is the main configuration panel of the application. For the results presented here, the groups were already pre-configured to make the  played games comparable.}
  \Description{Game configuration: setting up fairPlay using configurable parameters}
  \label{fig:gConfig}
\end{figure}

\subsection{Game Configuration} \label{Game Configuration}

% tag: "Q3" - first \added in below paragraph, dataset
% tag: "Q2" - second \added in below paragraph, p-value
Before entering the game, the configuration page allows players to select the specific dataset they want to work on, choose the machine learning algorithm to be applied, and express their preferences for certain population groups.
%each bearing a specific function and impact, 
as illustrated in Figure \ref{fig:gConfig}.\par
As previously noted, FairPlay primarily concentrates on pre-processing methods, targeting the dataset preparation stage before it is utilized in any training process. The configuration page enables users to select their desired dataset. In this study, a Hiring dataset was used, but the approach is readily applicable to other tabular datasets. The 'Hiring' dataset is a synthetic dataset that was originally introduced in D-BIAS \cite{ghai2022d}, designed to mimic a typical hiring scenario for controlled experimentation and analysis. It consists of 4,000 entries, each representing a fictional job candidate, with key data entry fields including age, gender, race, work experience, Grade Point Average (GPA), SAT score, college rank, major, and a binary hire decision variable. \added{We chose this particular synthetic dataset to seamlessly extend the D-BIAS system from single-player to multiplayer. It was well vetted and many experiments had been done with it.} Once the dataset is selected, its features are displayed for the user and a label (outcome) variable is chosen. The initial causal network, represented as a Directed Acyclic Graph (DAG), is inferred using the PC algorithm, a causal discovery method named after its creators Peter Spirtes and Clark Glymour \cite{spirtes2000causation}. 
% \textcolor{red}{I am not sure if this extensive discussion on p-values is really needed. Maybe just a shorter version. Rather, it should be acknowledged that the PC algorithm will not always return a perfect causal DAG which needs to be corrected by experts. I would prefer to state that the system reads in a correct DAG, created by experts with a system like D-BIAS or others beforehand. In that case you do not need Fig. 1 either since stakeholders get the graph already.}
This algorithm uses a p-value which is a threshold for statistical significance. More details on how the DAG is derived and the p-value's relevance is described later in section \ref{section:before-game}. The default p-value is set to 0.01, ensuring a high level of confidence in the results for most scenarios. Users can proceed with this default setting without needing a deep understanding of p-values. For those with specific requirements, the system allows adjustments to the p-value, assuming users understand the implications of such changes.

% It is important to note that the PC algorithm may not always return a perfect causal DAG. Expert validation and correction are often required. Therefore, The system can also read in a pre-validated DAG created by experts using tools like D-BIAS or other reliable methods, ensuring the accuracy and relevance of the causal model.
% \added{The default p-value is set to 0.01, Users who are not familiar with the concept of p-value can proceed with the default setting, as it is suitable for most scenarios and ensures a high level of confidence in the results. The selected p-value remains constant throughout a game session or user study, maintaining consistency in the analysis. For users who wish to adjust the p-value based on their specific requirements, the system allows them to do so. However, it is expected that users who choose to modify the default setting have a clear understanding of the concept and its implications. While the current version of FairPlay does not include an in-depth explanation of p-values, future iterations could incorporate additional resources or tooltips to provide users with more context and guidance when adjusting this parameter.}\par

% On the configuration page, users also have the option to pick from common ML algorithms like Logistic Regression, SVM, Naive Bayes, kNN, Decision Tree, or Neural Network. The chosen algorithm plays a role in continuously monitoring and logging classification performance metrics, providing valuable insights into how data debiasing impacts the model.\par

% tag: cML
On the configuration page, users also have the option to pick from common ML algorithms like Logistic Regression, SVM, Naive Bayes, kNN, Decision Tree, or Neural Network. The chosen algorithm plays a role in continuously monitoring and logging classification performance metrics, providing valuable insights into how data debiasing impacts the model. The purpose of these ML algorithms is to track standard performance metrics, which are computed at each stage of the data debiasing process. This approach allows us to measure the impact of debiasing on classification performance throughout the process \footnote{Technical details of the ML models, their implementation, and data usage can be found in Appendix \ref{appendix:mlmodels}.}.
% tag: "TGC"

 Players can select their role, a feature not available in D-BIAS, a single-user platform. Additionally, players specify a population group based on attributes they consider important. By clicking on the ‘Create Group’ button, a pop-up page appears for selecting features and their respective values for user preferences selection or group creation (see figure \ref{fig:Cgroup}).
 % \textcolor{red}{Not sure what these groups do later on.} 
 This involves specifying preferred features and value ranges for those features, aiding in the creation of group-based evaluations that assist in more informed decision-making throughout the game. While currently limited to one group per player, future updates could enable handling multiple groups or sub-groups.\par
A reset option is also available for restarting the game. By selecting "Enter Game", players move into the main game environment.

\subsection{The Game}

Upon entering the game with the chosen game configuration, the system 
reads in the initial causal network for the selected domain scenario and computes the initial game metrics for all players. The game features two main panels: the causal network view panel, with which players interact and manipulate, and the game metrics panel, which tracks and displays the game metrics to guide player decisions.

\begin{figure}[t!]
  \centering
  \includegraphics[width=\linewidth]{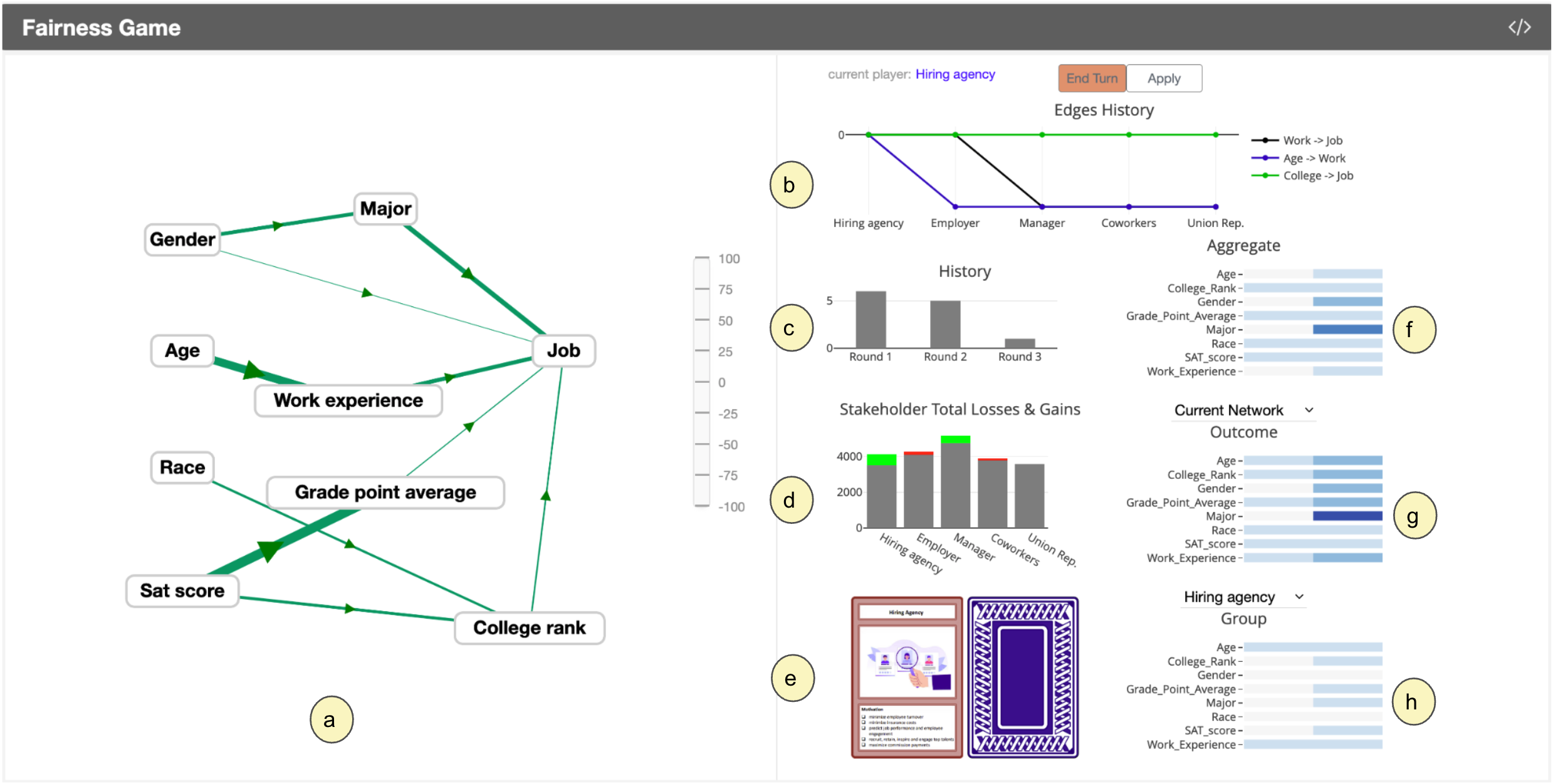}
  \caption{FairPlay Game Interface. The components are (a) causal network link editor, (b) edge history chart, (c) aggregate edge history chart, (d) stakeholder total loss and gain chart, (e) active stakeholder card stack, (f) aggregate attribute disparity chart, (g) attribute outcome chart,  (h) stakeholder attribute priority chart.}
  \Description{FairPlay: complete game dashboard}
  \label{fig:fullgame}
\end{figure}

\subsubsection{Causal Network Link Editor}

The central mechanism of the game resides in the causal network view, situated in the left panel (see figure \ref{fig:fullgame}). All features of the chosen dataset are represented as nodes in the network, and each edge represents a causal relation. The edge's width encodes the magnitude of the corresponding standardized beta regression coefficient which signifies the importance of the source node in predicting the target node, and the arrow indicates the direction of the causal relationship. For instance, in the Hiring dataset, the feature "Age" influences the feature "Work Experience", which directly affects the target node "Job". 

% \textcolor{red}{D-BIAS generates datasets by simulating data distributions via the causal network. As explained in \cite{ghai2022d}, when the weight of an edge in the causal diagram is decreased, it introduces more randomness into the corresponding antecedent node variable. For example, in figure \ref{fig:fullgame}(a), decreasing the edge weight of Gender to Major (where Gender=1 corresponds to males and Major = 1 corresponds to computer science majors) will result in a more balanced distribution of females majoring in CS, given all other qualities being equal, and subsequently gaining the Job. In this way, D-BIAS simulates a dataset that sheds the historical (male) bias in the originally collected data. Furthermore, by adding randomness instead of simply removing the sensitive Gender variable it also lowers proxy biases in variables downstream from them in the causal graph. }

A slight deviation from the base system D-BIAS, we have streamlined the causal edge operation in a manner that simplifies the gaming perspective. In this study, we have omitted additional edge operations, including adding, deleting, directing, and reversing causal edges, to focus on weight instead of topology. The default causal network is presumed to encompass all pertinent edge connections, and the weight of each edge can be adjusted within a range. By selecting an edge, players can change the edge weight by sliding the slider up or down, between -100 percent to +100 percent of its original weight \footnote{Details on how changing the edge weight affects the data are provided in Appendix \ref{appendix:AEPA}.}.
% Adjusting an edge weight by -100 percent effectively removes the influence of that feature on the connected node.  Conversely, adjusting an edge weight by +100 percent doubles the strength of the connection, which can amplify existing biases and disparities \cite{ghai2022d}.
% \textcolor{red}{of its original weight}. 
This bounded range of adjustments was chosen deliberately to maintain the coherence and balance of the gameplay, preventing extreme or unrealistic modifications that could disrupt the overall fairness dynamics of the game. By defining a reasonable range, users can focus on the relative impact of the edge weights rather than being overwhelmed by an entire numerical spectrum. Users need to find a relative balance within this scale to adjust the causal network appropriately. This range allows users to experiment with the strength of causal relationships while avoiding extremes that could either oversimplify the network or exacerbate bias.

\subsubsection{Game Metrics and Charts}

Each player's move depends on the current state of the causal network, group concerns, game metrics and score. This information is located on the right panel of the game interface in figure \ref{fig:fullgame}. At the top left, the current player is displayed. In the game, once a player adjusts the causal network and clicks "Apply", the game's metrics are updated, allowing the player to review these metrics and other players' scores before ending their turn with "End Turn". This two-stage process is designed to encourage active reflection on the actions taken. \added{This sequential, turn-based approach ensures that each player has an equal opportunity to modify the causal network without the gameplay descending into chaotic or uncontrollable behavior. In contrast, a simultaneous approach could result in conflicting changes, where multiple players attempt to adjust the network in ways that conflict or overlap, leading to a lack of order in the game flow.} \replaced{The pause for reviewing metrics allows players to carefully consider the impact of their changes and weigh potential trade-offs, thereby fostering thoughtful and deliberate decision-making. Additionally, once players click “Apply”, they are restricted from making further alterations, ensuring that every participant has a fair chance to contribute, maintaining a balanced and orderly progression of the game.} {The pause for reviewing metrics ensures players consider the impact of their changes and weigh any potential trade-offs, thereby promoting thoughtful and deliberate decision-making. Moreover, the game's design restricts players from making further alterations to the network once they've clicked "Apply". This rule is implemented to prevent players from continuously adjusting the causal network, ensuring that every participant gets a fair opportunity to play, maintaining a balanced and orderly flow of FairPlay.}

The edge history chart (see figure \ref{fig:fullgame} (b)), a line chart, tracks edge changes throughout the game, highlighting the top three edges with the highest percent change in edge weight. The X-axis indicates the player who made the change, while the Y-axis tracks the percent change in edge weight. If a specific edge is selected, the chart updates to show the history of that particular edge instead. This functionality ensures that all edge changes are accounted for, whether they are among the top three or not. The chart showcasing frequently changing edges not only aids in identifying conflicts and areas of disagreement among players but also directs their attention toward these conflicts, thereby expediting the process of reaching a consensus.

The Aggregate edge history chart (see figure \ref{fig:fullgame} (c)) displays the aggregate edge change count per round, indicating if the game is progressing towards a common consensus. In an optimal scenario, the total count of edge changes in the final round should be the lowest of all rounds, ideally reaching zero.

The stakeholder attribute priority chart (see figure \ref{fig:fullgame} (h)) reflects the priorities of the current player, determined by the selections they made on the game's configuration page, specifically the group they formed. For our study, we've simplified all features to a binary scale, assigning each a value of either 0 or 1; we found that this made choices clearer and easier to navigate. Based on the groups users have created, the priority chart will showcase players' level of care. The chart visually represents the extent to which players value each feature, based on the groups they have established. If both levels of a feature's bar are colored blue, it implies the player equally values each subgroup of the target variable. Conversely, if both levels of a feature's bar chart appear gray, it indicates that the player doesn't consider these attributes or features as central to their goals. A more comprehensive explanation of how these colors are assigned will be provided in Section 4.

The attribute outcome chart (see figure \ref{fig:fullgame} (g)), shows the number of individuals from each subgroup being hired based on the current causal network setup. It is a diverging heatmap with 11 color levels, ranging from red (lowest level), to gray (neutral), to dark blue (highest level). This chart indicates how many people from each subgroup were hired.

The aggregate attribute disparity chart (see figure \ref{fig:fullgame} (f)) shows the differences in hiring outcomes relative to the current player's desired outcome. This provides a measure of deviation between the actual outcome versus the player's preferred outcome (more details in Section 4).

Incorporating charts that display the group that each player cares about, the current state, and the difference between their desired and current status enables players to track their advancements, identify the areas that require further modifications, and make informed decisions accordingly. Making this kind of information available to players is crucial to a successful consensus reaching process \cite{innes2004consensus}. Moreover, the visual depiction of the difference between the desired and current status serves as a motivating factor, encouraging players to actively engage in the game and work towards narrowing the gaps.

The stakeholder total loss and gain chart (see Figure \ref{fig:fullgame} (d)) provides players with a simple and effective way to assess their performance in the game and compare it to others. By indicating increases in scores with green and decreases with red on top of the bars, the chart allows players to easily observe their progress and relative standing. This visual representation serves as a tool for players to track their overall performance and gain insights into how they are doing compared to their peers. Further information regarding the calculation of these values will be elaborated upon in Section 4.

Located at the bottom left, a stack of players' cards (see Figure \ref{fig:fullgame} (e)) grants users access to their role-specific general goals and objectives within the game, providing them with insights into the intended outcomes they strive to achieve in their respective roles. It should be mentioned that in actual scenarios, where players are genuine stakeholders with clear intentions, these cards are redundant and therefore not needed. However, in our user studies, volunteers were asked to assume specific roles 
% \footnote{For the final user study, we recruited individuals whose professions closely matched these roles.}.{\textcolor{red}{
\footnote{For the final user study we recruited individuals with professions that matched these roles to some extent.}. 
To assist these users in remembering their objectives while playing these roles, we incorporated these cards as a helpful reminder \footnote{The cards are displayed in Appendix \ref{appendix:cards}.}.

FairPlay design aims to empower players to actively address
bias, navigate conflicts, and work collaboratively toward reaching a consensus. Moving on from describing different aspects of the platform, the next section will discuss the methodology behind the calculation of values for the plots described in this section.

\begin{figure}[b!]
  \centering
  \includegraphics[width=\linewidth]{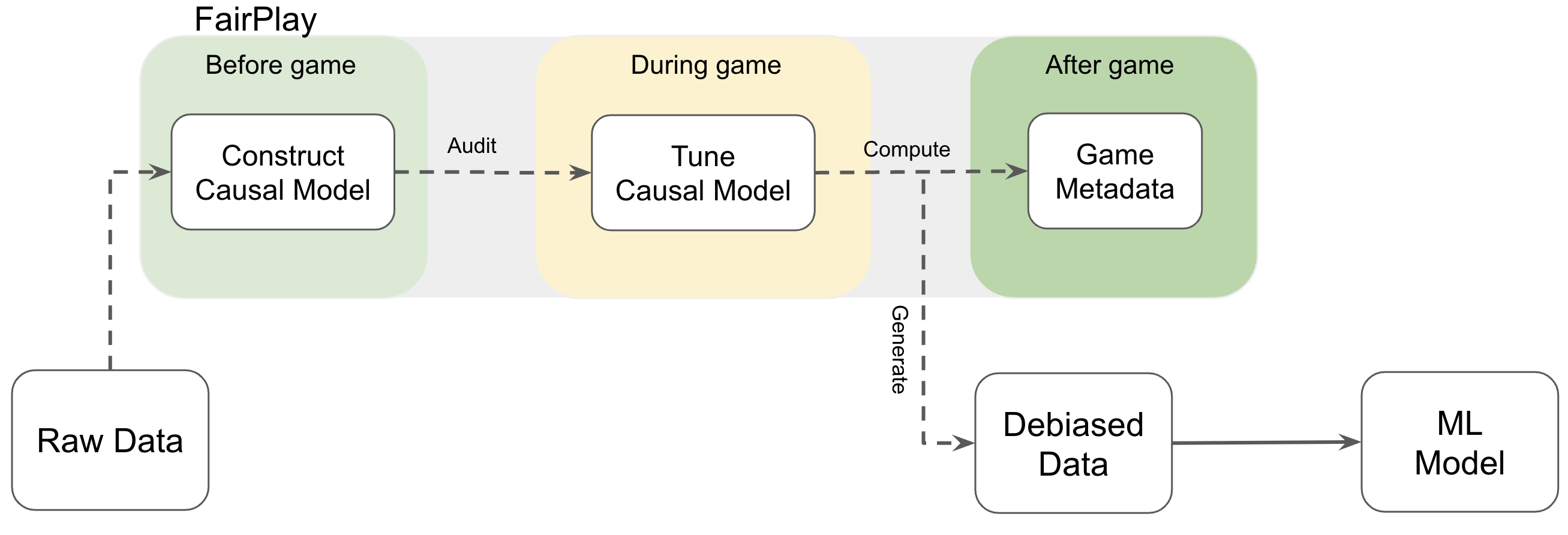}
  \caption{FairPlay System Overview. Detailed explanations are provided in the main text above.}
  \Description{Internal system components of FairPlay}
  \label{fig:sysOverview}
\end{figure}

\section{METHODOLOGY}
FairPlay is developed on the foundation of the previously published debiasing application, D-BIAS. In order to understand the system, we discuss its intricate aspects methodically. Figure \ref{fig:sysOverview} shows an overview. First, the default (initial) causal model is constructed from the Raw Data that would be used to train the ML model (first module in Figure \ref{fig:sysOverview}). The default outcomes are also displayed in the game interface. Then the game begins where the players seek to change the default outcomes per their priorities via iterative tuning of the causal model (center module in Figure \ref{fig:sysOverview}). The game ends when the players have achieved their goals which results in the Debiased Data. The Debiased Data can then be used to train any ML model. Here it is assumed that the ML model will not introduce biases on its own, else an independent ML model debiasing step would be required. The upper-right-most module of Figure \ref{fig:sysOverview} deals with analyzing user data produced during the game.

In the following we describe each of these three modules in detail. We begin by examining the technical aspects of the visual interface, followed by an exploration of the data storage for post-game analysis. 

\subsection{Before the Game: Game Setup} \label{section:before-game}
%In the previous section, we outlined the system's design. In this section, we examine the construction of each system element, elucidate the calculation of various scores or values, and explain the generation of charts and visual representations.    

%\subsubsection{Before Game}
Before the game begins, two crucial steps must be taken. Firstly, the construction of the causal network is required. Secondly, the players need to select the groups they care about based on their respective roles.

\paragraph{Generating Default Causal Model}

The authors of the D-BIAS paper provide a comprehensive explanation of the process used to generate the causal model, employing a widely-used causal discovery algorithm known as the PC algorithm \cite{colombo2014order}. The causal network is created utilizing the PC algorithm, which infers causal connections between variables based on conditional independence tests and orientation rules using the given p-value. Each node in the network symbolizes a data attribute, and the edges signify causal relations. Since automated causal inference can introduce incorrect or incorrectly directed edges, expert users would usually inspect the generated network and correct any errors. Therefore, our system can also read in a pre-validated DAG created by experts using tools like D-BIAS or other reliable methods, ensuring the accuracy and relevance of the causal model. For our studies, we have used the fully corrected model presented in the D-BIAS paper \cite{ghai2022d}.

The relationships between nodes are quantified using linear Structural Equation Models (SEM), which estimate the value of each node as a linear combination of its parent nodes. The regression coefficients in the model indicate the strength of causal relationships. Within this framework, a distinction is made between endogenous variables, nodes that have at least one edge leading into them, and exogenous variables, independent variables that have no parent nodes.

\paragraph{Creating Groups} \label{GroupCreate}
In the configuration page, players are required to create a group as explained in section \ref{Game Configuration}. A group is a set of prioritized attributes of the features, for example, for the GPA feature a player might prefer the high-GPA attribute. In the current game, it means that the player prefers that jobs are given to candidates with higher GPAs. Note that a player has a fixed budget of priorities. The more attributes the player selects the less priority is given to each. This ensures that players make thoughtful decisions about which attributes are most important to them.

\added{Although the preferences players set before the game are static and cannot be changed during gameplay, players may still need to adjust their goals as they negotiate and interact with each other. For example, while some of their preferred groups may not perform as well as they initially hoped (as indicated by red shades in the aggregate attribute disparity charts), players can compromise and agree that the final outcome is satisfactory enough to end the game. This process of adapting their goals within a fixed preference framework is key to reaching consensus, even when it means not all preferences are fully met (see Section \ref{Insights} for insights on goal adjustments).}

The stakeholder attribute priority chart (see Figure \ref{fig:fullgame} (h)) visually represents the selected and non-selected attributes for each variable, with the blue bars indicating the chosen attributes and the gray bars representing the non-selected ones. The objective of each player is to equally distribute their goal among the total selected attributes. For instance, if a player cares about 10 attributes, their level of concern for each attribute will be 10 percent (refer to Algorithm \ref{alg:gm}, Line 4-15), leading accordingly to lighter shades of blue for these attributes in the stakeholder attribute priority chart. We track and report various insights on groups for all the candidates using game analysis (see section \ref{Backdoor}).

\subsection{During the Game: Players Tune the Causal Model and Debias the Data }
During the game, the system aims to monitor modifications to the causal network, calculate metrics, and gather other game metrics. The computed metrics serve as valuable information for players, aiding them in making informed decisions for their next moves.
\paragraph{Tuning the Causal Model}
Once a player modifies the edge weight, we proceed to update the causal network with the new edge weight. Subsequently, we create a checkpoint of the updated causal network, labeling it as the ``\verb|Current Causal Network|''. Each new version of the updated causal network will be associated with this label, while older versions will be checkpointed for further analysis.

\paragraph{Debiased Dataset}
Whenever a modification is made to the causal network, the system generates a new dataset that differs slightly from the original data. The ultimate debiased dataset corresponds to the final causal network obtained after the game concludes. All intermediate datasets generated throughout the process leading to the final game stage are utilized for post-game analysis tracking. The algorithm for generating the debiased dataset is explained in the original D-BIAS paper \cite{ghai2022d} (Refer to Algorithm 1 in the original paper).

\paragraph{Computing the Game Metrics} \label{metrics}
The attribute outcome chart, Figure \ref{fig:fullgame}(g), offers a graphical display of the hiring outcomes according to the existing causal network at any point in the game. This illustrates the count of individuals from each subgroup whose label variable equals one, indicating in the present scenario that they have gotten the job. Players also have the ability to compare job status changes relative to both the current and original causal networks, allowing for a comprehensive assessment of the causal changes' impact on hiring outcomes (refer to Algorithm \ref{alg:gm}, Line 19-32). Two states of the outcome metric are being maintained, one for the current causal network and another for the original causal network. By tracking candidate groups and aggregating over the label (outcome) feature, corresponding to the "Job" feature in here, attribute-wise outcome metrics can be computed and presented in the aggregate chart, revealing deviations between actual and desired outcomes (refer to Algorithm \ref{alg:gm}, Line 34-42). These visualizations empower players to gain a deeper understanding of the game dynamics and make informed decisions to align their actions with their desired outcomes.

Another essential game metric is the total loss and gain, which represents the scores accumulated by each player throughout the game. This metric is computed by aggregating the sum of pair-wise multiplications between each player's group and the outcome (refer to Algorithm \ref{alg:gm}, Line 45-52).

%\subsubsection{Track Game Status}

%\subsubsection{Common battlefield for all players}

%\subsubsection{Causal Edge Strength} update => data

%\subsubsection{Computing Game Metrics} group/outcome/aggregate/tlag/gamepoint

%\subsubsection{Compute ML Metrics}

\subsection{After the Game: Data Collection and Game Analysis} \label{Backdoor}
In this section of the system overview, our primary emphasis lies on the post-game analysis and evaluation of the game. This phase involves scrutinizing several aspects, including the causal network changes, game moves, and conducting analysis. Further elaboration on these analyses will be provided in the upcoming section \ref{Results}.

% \subsubsection{Game Metadata} \label{metadate}

Every move the players make during the game is saved, and the data gathered opens room for extensive analysis after the game. This systematic tracking of changes in the causal network and each players outcomes allows for a comprehensive understanding of the evolving network and facilitates the evaluation of player interventions.

Another part of our post-game analysis includes ML metrics that play a crucial role in evaluating the performance of the machine learning algorithm. The algorithm used to assess these classification metrics is selected on the game configuration page (refer to Section \ref{Game Configuration} or Figure \ref{fig:gConfig}) and we track several standard ML metrics: Predicted Accuracy, representing the model's overall correctness. Predicted F1, a balanced measure of precision and recall. Individual Fairness, defined as the mean percentage of a data point’s
k-nearest neighbors that have a different output label, measuring equality and consistency in decision-making within a system. Parity, a metric used to assess equality in outcomes across demographic groups.

% Analyzing these metrics is crucial for assessing the impact of causal changes on the current debiased data in comparison to the original data. \added{To prevent information overload, we currently do not display these metrics to the players; instead, we track them internally for analysis purposes.} \deleted{In future work, we plan to incorporate these metrics as configurable parameters within the player's view, enabling a more informed decision-making process.}

% Game metrics play a crucial role in assessing the gameplay dynamics and the impact of player actions on bias removal. These metrics are computed using the current causal network, the original causal network, and the groups created by players; refer to section \ref{GroupCreate}.

% By incorporating these diverse game metrics, we can effectively analyze player actions, attribute distributions, hiring outcomes, and overall game progression. These metrics provide players with valuable insights and feedback, empowering them to make informed decisions and actively contribute to the removal of biases within the game's causal network.

Analyzing game metrics is crucial for assessing the impact of causal changes on the current debiased data compared to the original data. To prevent information overload, we currently track these metrics internally for analysis purposes, without displaying them to players. By incorporating these diverse metrics, we can effectively analyze the outcome of players' actions, final hiring decisions, and overall game progression.

\section{Experiments}

\subsection{Setup}
% To evaluate the effectiveness of FairPlay, we conducted 3 user studies, i.e., game play sessions. These user studies aimed to achieve several goals:

To evaluate the effectiveness of FairPlay, our research question aimed to determine whether consensus can be achieved among players in a multi-player game environment while modifying the causal graph to mitigate bias. The study goals were:

\begin{itemize}
  \item \textbf{G1}: Assess the game's ability to educate and engage players in the complexities of bias mitigation through their interactions and feedback during gameplay.
  \item \textbf{G2}: Gather insights into the consensus-building process within the game, observing how players collaboratively modify the causal graph and reach a consensus on removing bias and how they perceive the process and the outcome of FairPlay.
  \item \textbf{G3}: Analyze the outcome of the process, the debiased datasets, using accuracy and fairness metrics.
  \item \textbf{G4}: Collect feedback on the usability of the game interface and mechanics.
  \end{itemize}
% For the user studies, we recruited volunteers from the Computer Science graduate student population at a university. A total of 15 volunteers were sought for participation. Once we received responses from interested individuals, we randomly divided them into three groups of five participants each. The games were scheduled to be conducted remotely via Zoom. During the game sessions, players took turns and requested remote access to the host's screen to play their respective turns.

% For the user studies, we recruited volunteers online, from people affiliated with the Computer Science department at a major university. A total of 15 volunteers were sought for participation. On average, participants demonstrated familiarity with AI, ML, and fairness, along with a solid understanding of current issues in these domains. The participants were not limited to a specific demographic group.

For the user studies, we recruited volunteers online, from individuals affiliated with the Computer Science department at a major university. For the first three studies, this included employees and students. For the fourth study, we looked for participants with experience in industry for the specific roles we needed for the study.
A total of 20 volunteers were sought for participation. On average, participants demonstrated familiarity with AI, ML, and fairness, along with a solid understanding of current issues in these domains. The participants were not limited to a specific demographic group \added{with the majority being between 21-30 years old (41.7\%) and primarily male (75\%). Ethnically, the largest group identified as Asian (75\%). A significant portion of participants held a Master's degree (27.3\%), followed by those with Bachelor’s and Doctoral degrees.} More details about the demographic information of participants and their backgrounds is available in Appendix \ref{appendix:users}.

% Once we received responses from interested individuals, we randomly divided them into three groups of five participants each. The games were scheduled to be conducted remotely via Zoom. During the game sessions, players took turns and requested remote access to the host's screen to play their respective turns.

% tag: Qsync
Once we received responses from interested individuals, we randomly divided them into three groups of five participants each. Additionally, for the last user study, we recruited 5 participants separately, each with real-world experiences in a particular role we needed. The games were scheduled to be conducted remotely via Zoom, allowing for remote participation and not requiring players to be co-located. The game was played synchronously, with each player taking their turn to make changes to the causal network. During the game sessions, players took turns and requested remote access to the host’s screen to play their respective turns. Once a player ended their turn by clicking “End Turn,” the next player could make their move. This synchronous play style ensured that every participant could get a fair opportunity to play, maintaining a balanced and orderly flow of gameplay in which participants could observe other players' actions and understand their goals better.

In real-world applications, users would typically access the game through a web address on their own systems and would only be able to make changes to the game during their turn. However, for the user studies, we conducted sessions over Zoom to monitor user interactions and discussions. To maintain consistency across our first three studies, we predefined profile preferences and the groups each player cared about, rather than allowing users to make these selections themselves. We thoroughly explained these preferences to participants so they could act in line with their assigned roles. Player cards were used to help users keep track of their role-specific general goals. The roles selected for this particular dataset and user studies included Hiring Agency, Employer, Union Representative, Co-workers, and Manager. These roles represent stakeholders involved in debiasing training data for a job-applicant decision system in a real-world context as previously discussed. More information about these roles and each roles' goals and preferences is discussed in Appendix \ref{appendix:cards}. 

A fourth user study was conducted with participants recruited based on their experience in one of the aforementioned roles to better approximate real-world conditions. In this study, participants selected their own preferences based on their experience in that position before starting the game. This approach allowed us to assess our goals in a more realistic scenario and ensured that the observed results were not solely influenced by how we designed each role's preferences.

\subsection{GamePlay}

To familiarize participants with the FairPlay game mechanics, we began each user-study session with a ten-minute informational video. This video detailed how the game functions and how players could utilize its features to achieve their objectives. Following the video presentation, we conducted a brief question-and-answer session to address any queries participants had about the game.

Once all questions were addressed, the game commenced with each of the five players selecting a unique role. In a given round, players were tasked with modifying the weights in the causal diagram to align them with their respective goals. Upon satisfaction with their adjustments, players hit the 'Apply' button to review the results of their interventions. Subsequently, they ended their turn, allowing the next player to perform their modifications.

At the conclusion of a full round involving all five players, a popup displaying all players' scores was shown. Players were then asked whether a consensus to accept the current network had been reached or if they wished to continue modifying it. If even a single player opted to continue, the game extended into another round. %Alternatively, we can also gauge consensus when the amount of edge changes falls below a minimum. 

We conducted a total of four user studies. In the first three studies, the game continued for 5, 2, and 3 rounds respectively, lasting for 1 hour, 1.5 hours, and 1.5 hours. Our final study with participants with real-world experience continued for 3 rounds and lasted 1 hour. By the end of each game, all players achieved improved metrics reflective of their objectives. In the next section, we will discuss the results of these user studies.

\section{Results} \label{Results}

\subsection{Machine Learning Metrics Analysis} \label{MLMetrics}

Evaluating how classifiers perform on the final debiased datasets, particularly in terms of accuracy and fairness, is vital (\textbf{G3}). Accuracy is a primary concern in real-world applications, and having a fairer dataset is what the process aims to achieve. To evaluate the effectiveness, we consider 4 metrics detailed in section \ref{Backdoor} and displayed in Table \ref{tab:mlt}. Predicted Accuracy, reflecting the model's overall correctness, is conventionally sought at higher values; however, the debiased models reveal lower scores, signaling a deliberate trade-off for heightened fairness. Similarly, Predicted F1, a metric balancing precision and recall, is typically favored at higher values, yet the debiased models exhibit lower figures. Examining Individual Fairness, where lower scores indicate reduced disparate treatment among individuals, the debiased models consistently achieve significantly better (lower) values, indicative of a noteworthy improvement. Assessing Parity, a metric gauging equality in outcomes across demographic groups, higher values are preferred, and the debiased models generally exhibit enhanced parity values. These findings align with existing research on the tradeoff between accuracy and fairness \cite{pmlr-v81-menon18a}\cite{NEURIPS2019_b4189d9d}.

% New Table 1 - Include this

% \begin{table}[h]
% \centering
% \resizebox{\textwidth}{!}{
% \begin{tabular}{|c|c|c|c|c|c|c|c|c|}
% \hline
%  & \multicolumn{2}{c|}{User Study 1} & \multicolumn{2}{c|}{User Study 2} & \multicolumn{2}{c|}{User Study 3} & \multicolumn{2}{c|}{User Study 4} \\
% \cline{2-9}
%  & Original & \replaced{Debiased} & Original & \replaced{Debiased} & Original & \replaced{Debiased} & Original & \replaced{Debiased}{} \\
% \hline
% Predicted Accuracy & 0.76 & \color{red}{0.63} & 0.76 & \color{red}{0.69} & 0.76 & \color{red}{0.63} & 0.76 & \color{red}{0.63} \\
% \hline
% Predicted F1 & 0.58 & \color{red}{0.44} & 0.58 & \color{red}{0.53} & 0.58 & \color{red}{0.44} & 0.58 & \color{red}{0.44} \\
% \hline
% Individual Fairness & 22.11 & \color{green}{0.28} & 22.11 & \color{green}{\textbf{3.21}} & 22.11 & \color{green}{1.73} & 22.11 & \color{green}{0.04} \\
% \hline
% Parity & 35.96 & \color{green}{\textbf{47.45}} & 35.96 & \color{green}{45.9} & 35.96 & \color{green}{45.86} & 35.96 & \color{green}{\textbf{47.69}} \\
% \hline
% % False Positive Rate & 66.78 & 0.38 & 66.78 & 5.72 & 66.78 & 3.45 & 0 & 0 \\
% % \hline
% % False Negative Rate & 58.5 & 0 & 58.5 & 0 & 58.5 & 0 & 0 & 0 \\
% % \hline
% % Accuracy & 39.3 & 0.2 & 39.3 & 4.45 & 39.3 & 1.88 & 0 & 0 \\
% % \hline
% \end{tabular}
% }
% \caption{ML Metrics observed during all four user studies}
% \label{tab:mlt}
% \end{table}

\begin{table}[h]
\centering
\resizebox{\textwidth}{!}{
\begin{tabular}{|c|c|c|c|c|c|c|c|c|}
\hline
 & \multicolumn{2}{c|}{User Study 1} & \multicolumn{2}{c|}{User Study 2} & \multicolumn{2}{c|}{User Study 3} & \multicolumn{2}{c|}{User Study 4} \\
\cline{2-9}
 & Original & Debiased & Original & Debiased & Original & Debiased & Original & Debiased \\
\hline
Predicted Accuracy & 0.76 & 0.63 \reddown & 0.76 & 0.69 \reddown & 0.76 & 0.63 \reddown & 0.76 & 0.63 \reddown \\
\hline
Predicted F1 & 0.58 & 0.44 \reddown & 0.58 & 0.53 \reddown & 0.58 & 0.44 \reddown & 0.58 & 0.44 \reddown \\
\hline
Individual Fairness & 22.11 & 0.28 \greendown & 22.11 & 3.21 \greendown & 22.11 & 1.73 \greendown & 22.11 & 0.04 \greendown \\
\hline
Parity & 35.96 & 47.45 \greenup & 35.96 & 45.9 \greenup & 35.96 & 45.86 \greenup & 35.96 & 47.69 \greenup \\
\hline
\end{tabular}
}
\caption{ML Metrics observed during all four user studies. The green color indicates improvement, and the direction of the triangles shows how the value changed. For example, a green triangle pointing down means the value decreased, and lower values are preferred for this feature, so this decrease represents an improvement.}
\label{tab:mlt}
\end{table}

\subsection{Behavioral Observations} \label{Behavior}

Our initial analysis of whether players perceived the final outcome as fair (\textbf{G2}) was assessed through a question in our post-game survey. Participants were asked to rate their agreement with the statement: "I think that the activities led to a fairer system." on a scale from 1 (strongly disagree) to 5 (strongly agree). On average, users rated this question 3.7, suggesting that they believed the collaboration had a positive impact.

Also, throughout the user studies, participants were encouraged to vocalize their thoughts while playing, discussing the factors influencing their decisions each round. With their consent, the studies were recorded for more in-depth analysis later on. This was to be able to analyze the consensus-building process in FairPlay in more detail (\textbf{G2}).

Qualitative analysis of players' dialogues throughout the game helped us assess whether the dashboard features were assisting or confusing them. The dashboard appeared intuitive to players, even those with no prior experience with causal networks. Players modified edges based on attributes they were supposed to care about, stating things like, "I'm doing this because I care about feature X." They also adjusted edges previously edited by others, saying, "I don't care about this feature, so I don't want this to play a role." Additionally, players predicted how their changes would affect the groups they cared about, with statements such as, "I want more people with feature X to get the job." In most cases, their predictions aligned with the results shown in the right panel plots after clicking the "Apply" button, indicating that they were able to use the causal network correctly to achieve their desired outcomes.

We can analyze user behaviors when it comes to the right panel and how insightful it was in the game through the lens of two different philosophical schools of thought: Consequentialism (which focuses on outcomes) and Deontology (which focuses on the morality of actions) \cite{alexander2007deontological}. Prior to the studies, all participants were asked whether their ethical approach aligned more closely with Deontological or Consequentialist principles. 72.7\% of participants identified as Deontologists, while 27.3\% identified as Consequentialists. This distinction in mindsets was evident in the way users made their decisions. Some players adjusted the network to achieve the best outcome metrics for their groups, while others prioritized setting the network parameters correctly, regardless of the outcomes shown by the plots. The Deontologist players showed a strong preference for down-weighting edges that emerged from the sensitive variables \footnote{Many variables are explicitly defined as “sensitive” by specific legal frameworks \cite{10.1145/3616865}. In our dataset, Age, Gender, and Race are considered sensitive based on the framework outlined in \cite{sensitive-variables}.}, regardless of the consequences of their metric outcomes, hence not paying too much attention to the right panel. If we consider only the Consequentialist players, we notice they were more likely to fiddle with parameters in either direction while searching for the best outcome metrics. They would first look at the panel on the right to see how their groups are doing, modify edges accordingly, and then observe the outcomes on the right panel more thoroughly.

Despite these differences between strategies, a consensus was eventually reached in all user studies, demonstrating the game's ability to facilitate mutual agreement even among diverse objectives by providing intuitive means of modifying the network and insightful metrics to help users make decisions (\textbf{G2}).

To determine if the game educated players about the complexities of bias mitigation (\textbf{G1}), we asked users to rate the following statement in our post-game survey on a scale from 1 (strongly disagree) to 5 (strongly agree): "The game improved my understanding of fairness and bias in automated decision systems." The average score was 3.3, indicating that the game had an overall positive effect on educating the players.

\subsection{System Usability Score Analysis} \label{Feedbackform}

\begin{figure}[h]
  \centering
  \includegraphics[width=\linewidth]{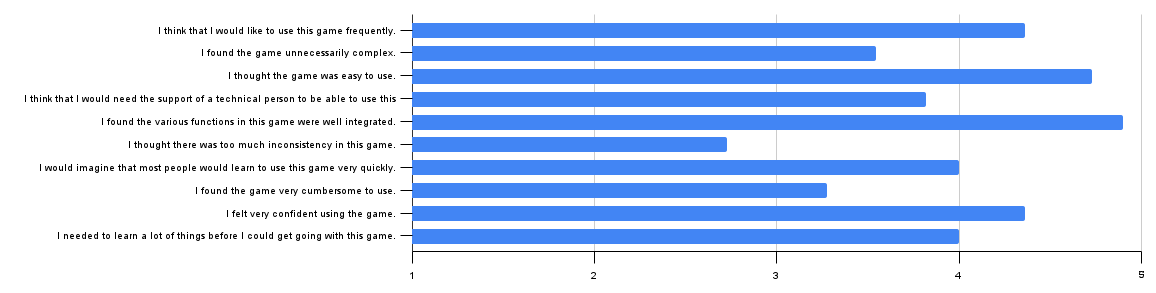}
  \caption{FairPlay Feedback Results}
  \Description{FairPlay Feedback Results}
  \label{fig:feedback}
\end{figure}

One of the key indicators of a tool's success, irrespective of its features and objectives, is its perceived effectiveness, efficiency, and user satisfaction (\textbf{G4}). To assess this, we utilized the standard System Usability Scale \cite{brooke1996sus} (created by John Brook at Digital Equipment Corporation in 1986), which employs a 5-point Likert scale. After completing the study, users were requested to fill out a feedback form. The user feedback statistics, as shown in Figure \ref{fig:feedback}, reveal that the statement players disagreed with the most was "I thought there was too much inconsistency in FairPlay", while the statement the players agreed with the most was "I found the various functions in FairPlay were well integrated." This reflects that the tool's visual presentations and functionalities were cohesively aligned and user-friendly. The overall SUS score was 68.05, positioning FairPlay as a positive and intuitive system, especially since SUS scores above 68 are considered above average.

% \subsection{Post-Game Analysis} \label{Backdoor} 

%\begin{figure}[h]
%  \centering
%5  \includegraphics[width=0.8\linewidth]{figures/us1.png}
 % \caption{FairPlay User Study 1 : Edge Weights vs. Round}
 % \Description{FairPlay User Study 1 : Edge Weights vs. Round}
 % \label{fig:feedback1}
%\end{figure}

%\begin{figure}[h]
%  \centering
%  \includegraphics[width=0.8\linewidth]%{figures/us2.png}
%  \caption{FairPlay User Study 2 : Edge Weights vs. Round}
%  \Description{FairPlay User Study 2 : Edge Weights vs. Round}
%  \label{fig:feedback2}
%\end{figure}

%\begin{figure}[h]
%  \centering
%  \includegraphics[width=0.8\linewidth]{figures/us3.png}
%  \caption{FairPlay User Study 3 : Edge Weights vs. Round}
%  \Description{FairPlay User Study 3 : Edge Weights vs. Round}
%  \label{fig:feedback3}
%\end{figure}

% contains Errors
% More detailed inquiry into the specific values showed that Gender$\rightarrow$Major, Gender→Job, and Age$\rightarrow$WorkExperience, were typically highly downweighted to zero or near-zero by the final round.} Furthermore, in User Study 3, by the final round almost all edges were zeroed out apart from Age$\rightarrow$WorkExperience and WorkExperience$\rightarrow$Job.

\subsection{Insights} \label{Insights}
In this section, we discuss the specifics of the four user studies and examine the outcomes of each game upon conclusion.

\begin{figure}
    \centering
    \includegraphics[width=\textwidth]{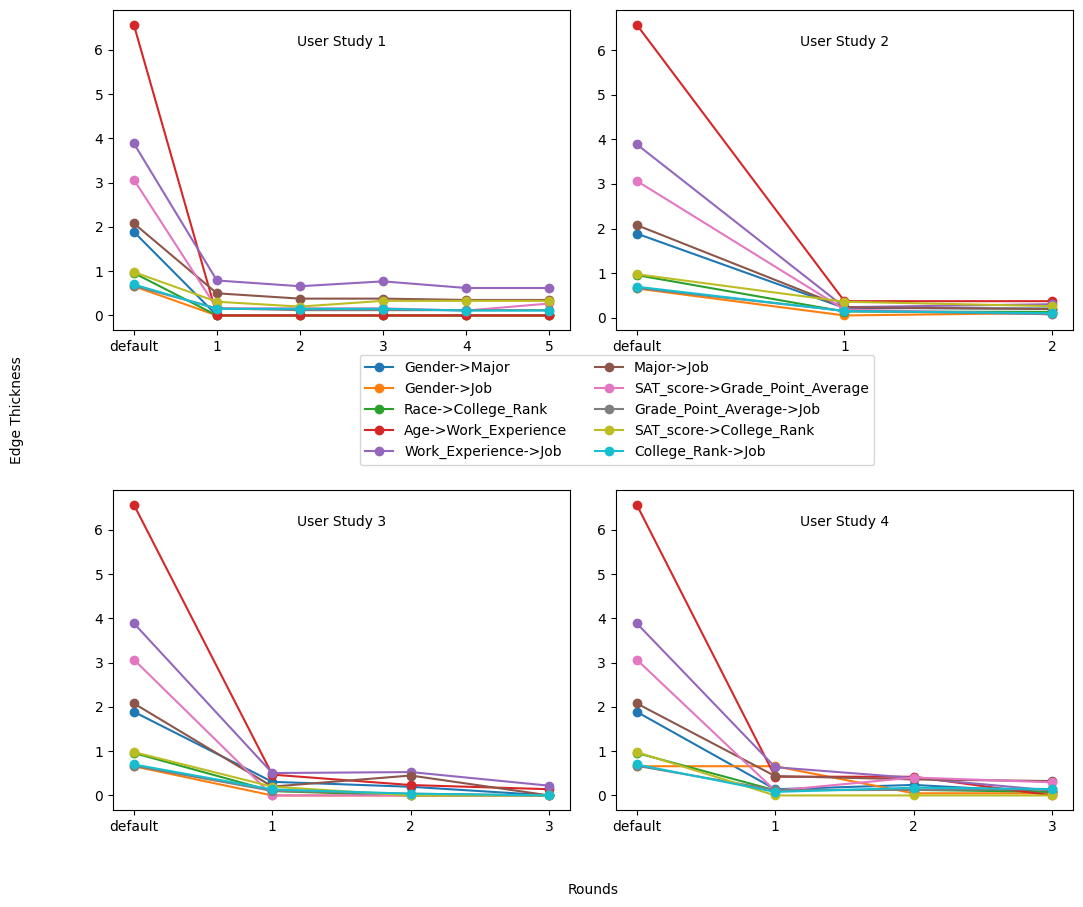}
    \caption{FairPlay User Studies: Edge Weights vs. Round for all User Studies}
    \label{fig:round-plots}
\end{figure}

Figure \ref{fig:round-plots} shows the progression of edge-weight adjustments made by players in each round.\\

% \begin{figure}
%      \centering
%      \subfloat[User Study 1]{
%         \includegraphics[width=0.45\textwidth]{figures/us1.png}
%         \label{fig:round-plots1}
%      }
%     \subfloat[User Study 2]{
%          \includegraphics[width=0.45\textwidth]{figures/us2.png}
%          \label{fig:round-plots2}
%     }\par
%          \subfloat[User Study 3]{
%         \includegraphics[width=0.45\textwidth]{figures/us3.png}
%         \label{fig:round-plots3}
%      }
%      \subfloat[User Study 3]{
%         \includegraphics[width=0.45\textwidth]{figures/us4.png}
%         \label{fig:round-plots3}
%      }
%     \caption{FairPlay User Studies: Edge Weights vs. Round for all three User Studies}
%     \label{fig:round-plots}
% \end{figure}

% \added{We note that like in many real-world scenarios and complex systems, we expect sensitive dependence on initial conditions. If the roles had been even slightly different, and tensions between roles more or less, we might expect largely different outcomes. As an extreme example, had we set up two roles to dispute sole-ownership of a non-fungible commodity, then we would have no guarantee of convergence. We also note that complex system can (and often do) have attractors. We did perform several different runs, all with different participants, who interpreted their goals differently. Despite this variation, patterns did emerge. Conservatively, we can say that we observed some measure of robustness in the system.}

In all four studies, there's a sharp decline in edge thickness from the default to the first round. This suggests that players were quick to act on their initial assessments of the causal network. Furthermore, nearly all edge-weights dropped below 1 in the opening round and remained low, hinting at the strength of initial user impressions.

By the second round and onwards, changes become less drastic, and edge weights appear to stabilize. This could indicate that users reached some form of consensus or satisfaction with the state of the causal network early on.

While the general trend across rounds is similar, the final edge weights vary between studies, suggesting that while the process is consistent, the outcomes are subjective and influenced by the unique dynamics and decisions of the participants in each user study.

\begin{figure}[h!]
  \centering
  \includegraphics[width=\linewidth]{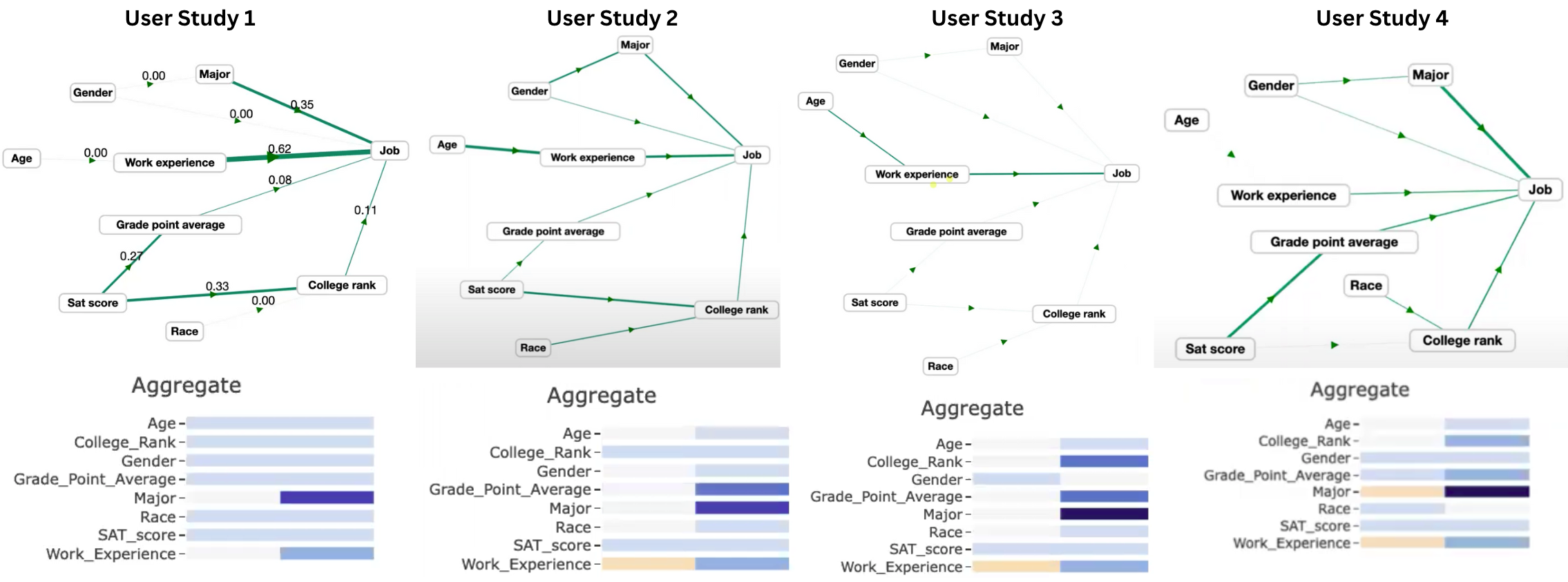}
  \caption{FairPlay Matrix. This figure compares the causal networks created by each set of players (top row) and the aggregate attribute disparity charts (bottom row)}
  \Description{FairPlay Matrix. . }
  \label{fig:matrix}
\end{figure}

%Another pattern of note is that WorkExperience$\rightarrow$Job is typically amongst the highest weighted edges in the final round. More detailed inquiry into the specific values showed that edges like Gender$\rightarrow$Major, and Gender$\rightarrow$Job were typically highly downweighted to zero or near-zero by the final round. Furthermore, in User Study 3, almost all edges were zeroed out by the final round.
Figure \ref{fig:matrix} displays the final causal networks and the aggregate attribute disparity charts at the conclusion of the gameplay.

Upon examining the final causal networks, a consistent pattern becomes evident across all four games. We observe sparse networks with many edges reduced to minimal weights, particularly for sensitive attributes like Age, Race, and Gender. This pruning of dependencies results in simple network topologies.

When we turn to aggregate attribute disparity charts, it's evident that players managed to avoid unfavorable outcomes (marked by red shades) for almost all attributes across all games.  However, exceptions are observed in Games 2, 3 and 4, concerning individuals with low work experience. This finding suggests that players accepted that individuals with less work experience may not be selected for the job, despite at least one participant valuing this attribute (otherwise it wouldn’t be displayed in a red shade). This shows that users are willing to accept trade-offs in order to reach consensus (The same holds true for Major in the forth study).

Drawing insights from Figure \ref{fig:matrix} as a whole, players seem to be striving for maximum scores for their groups, as indicated by the aggregate attribute disparity charts. Interestingly, it appears possible to optimize benefits for the desired groups for everyone without explicitly taking any particular definition of fairness into account. This demonstrates that although satisfying all definitions of fairness might be challenging, as suggested by the literature, reaching a consensus on what factors should influence the final decision in a specific context is achievable. Users were able to negotiate and agree on trade-offs, indicating a collective prioritization of certain attributes over others to reach a shared understanding of fairness within the specific context of the game.

\section{Discussion} \label{Discussion}
Our platform's objective was to create a tool that facilitates users in collaboratively determining the causal structure of their datasets. It's important to recognize that in the absence of our tool, attaining this goal would be a significant challenge. An alternative might involve engaging stakeholders in a dialogue to debate the causal structure. However, without the ability to modify the causal network and observe its impact on the data, and lacking a structured negotiation process, reaching a consensus is not only improbable but also likely of low quality due to the lack of informed decision-making. Our interface addresses these challenges by 1) Establishing a systematic approach to the negotiation process, 2) Ensuring that every participant's perspective is heard and considered, and 3) Providing users with sufficient information to make well-informed decisions. \par

% \added{Our approach of allowing continued changes as long as anyone was unsatisfied, amounted to using strict consensus. This approach forced users to reconcile all problems instead of skirting over some. While this did mean the process could take longer, it meant the final results were more robustly acceptable to the participants.}

As mentioned in section \ref{sec:relatedwork-CRP}, soft consensus tries to achieve two important goals: (i) Reflect better partial agreement; and (ii) Guide the Consensus Reaching Process (CRP) until an acceptable high level of agreement is achieved among the decision-makers \cite{ZHANG2019580}. Across all four user studies, stakeholders expressed verbal satisfaction with the existing graph. For some participants, this contentment emerged as early as the second round and persisted throughout the game, while for others, satisfaction fluctuated from one round to the next due to modifications made by other players. A crucial point to note is that the game was only concluded when every stakeholder expressed complete satisfaction with the current state. This condition was met in all four user studies, with players verbally confirming their satisfaction with the outcomes. This outcome serves as a clear confirmation of the CRP's success, demonstrating its effectiveness in achieving a high level of consensus among all participants.

\section{Limitations} \label{Limitations}

% This study identifies three primary limitations in the FairPlay software system, each of which merits careful consideration.

% \subsection{Player Cards} \label{Player Cards}

% The current software system is overly specialized for a specific set of roles, notably evident during gameplay where player cards, representing each role, are displayed. This contrasts with the game setup mode, where users have the flexibility to configure the game with any number of roles. The use of player cards with predefined roles restricts the system's general applicability and versatility. Therefore, it is advisable to eliminate this unnecessary specialization in future versions to enhance the system's adaptability and broad utility.

% \subsection{Learning Curve and User Engagement} \label{Learning Curve and User Engagement}
Our approach is subject to certain limitations that are discussed below.\\
% \added{Zeroth, while our system is called FairPlay, and it is an extension of a previous work called D-BIAS, this may be misleading.  Our system is grounded on the concept of preference optimization \cite{yudkowsky2004coherent} , as popularized by AI safety researcher Eliezer Yudkowsky.  What constitutes a sensitive variable, or whether that is an appropriate decision criteria, is seen as a preference.  Instead of decreasing bias, or increasing fairness, our system can be thought of as AI-alignment.  We align the bias of the system to the bias of the users.  Nothing in the literature suggests that anything else is possible.}
Firstly, the participants in our user studies were not actual stakeholders in a real scenario, which could influence the game dynamics. In real-world settings, stakeholders might approach the game with greater eagerness and potentially be less willing to compromise. Similarly, time-pressure may be less (or more) of an issue in real-world settings. Furthermore, since the sessions were conducted with participants aware that they were being observed, there's a risk that their behavior was influenced by the Hawthorne effect \cite{landsberger1958mla} (i.e. studying an agent, changes the agent's behavior). Although this was a conscious choice to facilitate observation of user interaction with the system, it remains a limitation of the study.

Another notable limitation was the learning curve associated with the system. Even with thorough training, participants initially struggled to use the software effectively to pursue their objectives. This challenge was most apparent in the first round, characterized by more experimental than tactical behavior. Despite this, the System Usability Scale (SUS) feedback indicated that users did not perceive the learning curve as steep. Nonetheless, to improve the process, enhancing the introductory phase of the game, potentially with a more interactive approach than a video, could be beneficial.

Lastly, we have deliberately simplified the information available to users. While this decision was made to prevent user overwhelm and aim for simplicity, it may pose limitations when scaling to real-world scenarios. The absence of machine learning metrics in the interface and the full range of edge operations for example, might impede stakeholders in a practical context. Addressing these constraints will be crucial for the interface's applicability in actual stakeholder environments.

\section{Future Work} \label{Future Work}

While FairPlay is primarily designed for practitioners and stakeholders across various domains, its utility is not limited to these groups alone. It also holds potential for application in other areas, such as educational settings. Current computer science courses addressing AI fairness and bias typically lean towards statistical analysis or incorporate philosophical perspectives that often lack pragmatic implications for students \cite{10.1145/3491101.3503568}. However, pedagogical research has long emphasized the benefits of employing tools and visualizations in enhancing student learning \cite{10.1145/2168931.2168945}\cite{ToolsForTeaching}. In this context, case studies have highlighted the positive effects of using publicly available visualization tools from Human-Computer Interaction (HCI) practice as educational resources to explore algorithmic fairness concepts \cite{10.1145/3491101.3503568}. Given its collaborative nature and the unique way it engages users, FairPlay emerges as an excellent resource for educational settings. Exploring the potential of FairPlay in aiding students to effectively engage with the identification and mitigation of bias presents a promising research direction. \par
FairPlay meets many of the design criteria for a fairness toolkit as recommended by practitioners \cite{10.1145/3411764.3445604}, yet there is scope for enhancement. Acknowledging the flexibility that practitioners often have at different stages of their machine learning pipelines \cite{10.1145/3411764.3445604}, it becomes apparent that FairPlay has the potential to expand its impact. By extending its collaborative approach to encompass various phases of the machine learning lifecycle, not just limited to the pre-processing stage, FairPlay could significantly enhance its utility and effectiveness. This could include allowing users to have their own ML models incorporated in the system where right now only a limited number of options are available. Moreover, while FairPlay currently supports only edge weight modification, we plan to introduce additional edge operations, such as deleting or adding edges, to provide players with greater flexibility. This will allow us to explore and analyze the impact of these operations on the consensus-reaching process. Another intentional design choice was to hide ML-related performance metrics from players to avoid overwhelming them with excessive plots and data. However, we aim to tackle the challenge of integrating these metrics into the player’s view in a manner that enhances informed decision-making without causing information overload in future work. \added{In the future, we also plan to allow users to specify more detailed preferences, such as age ranges (e.g., 25-35) and additional categorical preferences (e.g., education level) instead of just binary ones.} \par 
As highlighted in the related works section, it's clear that no algorithm can be solely relied upon to determine the most fitting definition of fairness or to establish an agreeable balance among various fairness metrics for all stakeholders. Consequently, FairPlay adopts a human-centered approach. Nevertheless, the idea of integrating an automated agent within FairPlay to assist in the consensus-building process among stakeholders is an intriguing concept. There are well-established consensus-reaching algorithms that could be adapted for such an agent \cite{Bhardwaj2020}\cite{TANG2021632}\cite{LIAO2016274}. Investigating how the inclusion of an automated agent might alter the group dynamics, and whether it aids the process, would be a valuable area for future research. This represents an innovative way of exploring the interaction between humans and algorithms in the context of addressing bias, offering potential insights into enhancing collaborative decision-making. \par

Moreover, we recognize the importance of making FairPlay accessible for replication, collaboration, and improvement by the broader community. We plan to develop FairPlay as an open-source project with detailed contribution guidelines and coding standards. This will facilitate collaboration, allowing researchers and practitioners to enhance functionalities and adapt the tool for various use-cases. \par
The promising outcomes FairPlay demonstrates in addressing fairness issues through collaboration suggest that this method could be beneficially adopted by other fairness toolkits as well. This would provide a diverse group of stakeholders with a structured framework for achieving consensus on complex and often contentious issues like fairness.

\section{Conclusion}

With the increasing reliance on algorithms as decision-makers across various contexts, the importance of thoroughly auditing these algorithms for ethical concerns has become more pronounced. However, research indicates that fulfilling all fairness criteria can be challenging, if not impossible. This necessitates a context-specific audit, ideally conducted by humans, though it is important to acknowledge that humans too have their own biases and blind spots. Consequently, adopting a team-based approach to this audit process emerges as a promising strategy. FairPlay aims to facilitate such an environment, where stakeholders or domain experts, ideally representing a diverse array of viewpoints, collaborate systematically to discern the relevant features in the underlying dataset. The varied outcomes and distinct final causal structures resulting from the four user studies underscore the necessity of tools like FairPlay that facilitate such processes. This diversity highlights that different groups may converge on varying agreements, leading to unique final structures. The fact that all four user studies achieved consensus serves as a strong validation of FairPlay's effectiveness. Developing tools like FairPlay is crucial for enabling informed auditing processes. Without such platforms, it would be unfeasible to engage in meaningful conversations that lead to prompt action, highlighting the crucial role these tools play in facilitating collaborative decision-making.

%%
%% The acknowledgments section is defined using the "acks" environment
%% (and NOT an unnumbered section). This ensures the proper
%% identification of the section in the article metadata, and the
%% consistent spelling of the heading.
\begin{acks}
Work reported in this publication was supported by SUNY System Administration under SUNY Research Seed Grant Award 23-01-RSG. ChatGPT was utilized to generate sections of this Work, including text, tables, graphs, code, data, citations, etc.
\end{acks}

\clearpage
%%
%% The next two lines define the bibliography style to be used, and
%% the bibliography file.
% \bibliographystyle{ACM-Reference-Format}
% \bibliography{sample-base}

%%%%%%%%%%%%%%%%%%%%%%%%%%%%%%%%%%%%%%%%
%% Compiled Bib - Start   %%%

%%% -*-BibTeX-*-
%%% Do NOT edit. File created by BibTeX with style
%%% ACM-Reference-Format-Journals [18-Jan-2012].

%%% Compiled Bib End
%%%%%%%%%%%%%%%%%%%%%%%%%%%%%%%%%%%%%%%

%%
%% If your work has an appendix, this is the place to put it.
\clearpage % Ensure new section starts on a new page

\renewcommand{\thesubsection}{\Alph{subsection}}
\appendix

\section*{Appendix}
\addcontentsline{toc}{section}{Appendix}
This appendix contains additional details and figures referenced in the main text. The following sections provide further insights into the experiments conducted, the configuration settings used, and technical details about machine learning models.

\subsection{Machine Learning Models} \label{appendix:mlmodels}
\addcontentsline{toc}{subsection}{Machine Learning Models}
% \label{appendix:mlmodels}

The ML models used in our study are implemented using the `scikit\-learn` library in Python. Below are the details of each model and the relevant code snippets for their implementation. The original data (`df`) and debiased data (`df\_deb`) are used to compute performance metrics at each stage when changes are made to the causal network by a player.

\subsubsection*{Logistic Regression}
Logistic Regression is a linear model used for binary classification tasks. It models the probability that a given input belongs to a certain class.

\begin{verbatim}
from sklearn.linear_model import LogisticRegression
model = LogisticRegression()
model.fit(X_train, y_train)
predictions = model.predict(X_test)
\end{verbatim}

\subsubsection*{Support Vector Machine (SVM)}
SVM is a supervised learning model that can be used for both classification and regression tasks. It works by finding the hyperplane that best separates the classes.

\begin{verbatim}
from sklearn.svm import SV
model = SVC()
model.fit(X_train, y_train)
predictions = model.predict(X_test)
\end{verbatim}

\subsubsection*{Naive Bayes}
Naive Bayes is a probabilistic classifier based on applying Bayes' theorem with strong (naive) independence assumptions between the features.

\begin{verbatim}
from sklearn.naive_bayes import GaussianNB
model = GaussianNB()
model.fit(X_train, y_train)
predictions = model.predict(X_test)
\end{verbatim}

\subsubsection*{k-Nearest Neighbors (kNN)}
kNN is a simple, instance-based learning algorithm used for classification and regression. It predicts the class of a sample based on the majority class among its k nearest neighbors.

\begin{verbatim}
from sklearn.neighbors import KNeighborsClassifier
model = KNeighborsClassifier(n_neighbors=5)
model.fit(X_train, y_train)
predictions = model.predict(X_test)
\end{verbatim}

\subsubsection*{Decision Tree}
Decision Tree is a non-parametric supervised learning method used for classification and regression. It splits the data into subsets based on the value of input features.

\begin{verbatim}
from sklearn.tree import DecisionTreeClassifier
model = DecisionTreeClassifier()
model.fit(X_train, y_train)
predictions = model.predict(X_test)
\end{verbatim}

\subsubsection*{Neural Network}
Neural Network is a set of algorithms, modeled loosely after the human brain, that are designed to recognize patterns.

\begin{verbatim}
from sklearn.neural_network import MLPClassifier
model = MLPClassifier()
model.fit(X_train, y_train)
predictions = model.predict(X_test)
\end{verbatim}
\subsubsection*{Data Usage}
The original data (`df`) and the debiased data (`df\_deb`) are used to evaluate the performance of the ML models. The debiased data is updated at each stage when a player makes changes to the causal network. The performance metrics are computed by comparing predictions on the original data and the debiased data. Here is an example of how the data is used:

% \begingroup
% \setlength{\parskip}{5pt} % Reduce space around verbatim
% \setlength{\parindent}{5pt}
% \vspace{-\topsep}
\begin{verbatim}
# Original data
X_train, X_test, y_train, y_test = train_test_split(
    df.drop(columns=['label']), df['label'])
# Debiased data (updated after each change in causal network)
X_train_deb, X_test_deb, y_train_deb, y_test_deb = train_test_split(
    df_deb.drop(columns=['label']), df_deb['label'])
# Model training and evaluation
model = LogisticRegression()
model.fit(X_train_deb, y_train_deb)
predictions = model.predict(X_test_deb)
# Compute performance metrics
accuracy = accuracy_score(y_test_deb, predictions)
\end{verbatim}
% \vspace{-\topsep}
% \endgroup

\vspace{1em} % Add some vertical space after the code snippet

These code snippets demonstrate the standard implementation of the ML models using `scikit-learn` to compute the performance metrics at each stage of the data debiasing process.

% \clearpage % Ensure new section starts on a new page

\subsection{Additional Figures}
\addcontentsline{toc}{subsection}{Additional Figures}
This appendix includes additional figures referenced in the main text.

% \vspace{-5em} % Reduce space before the figure
\begin{figure}[H]
  \centering
  \includegraphics[width=0.4\linewidth]{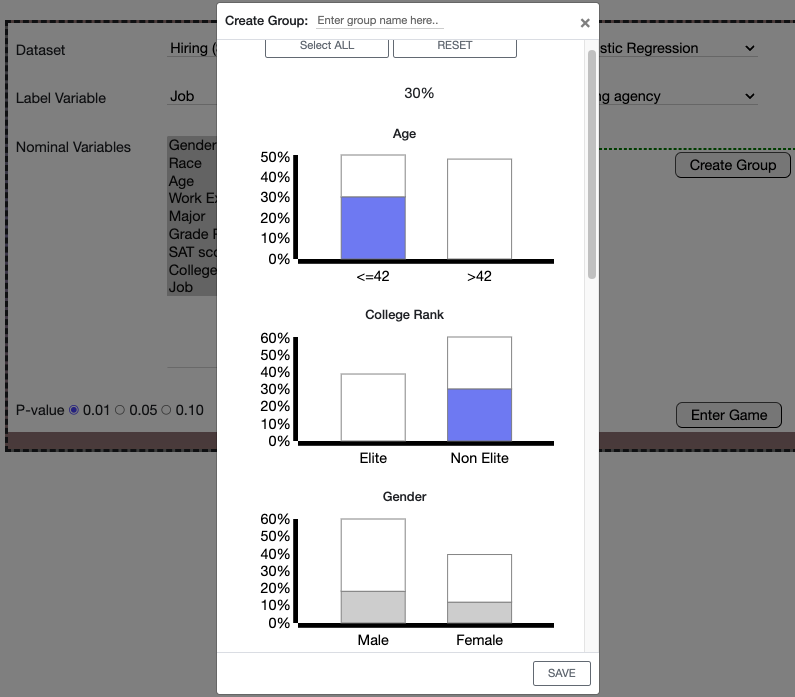}
  \caption{Game Configuration: User Preferences Selection or Group Creation. The height of each rectangle indicates the percentage availability of the corresponding attribute in the dataset. Selected attributes’ rectangles will be filled with blue, partially or fully, depending on the percentage of the attribute’s inclusion in the group. Unselected attributes are filled with gray, indicating their availability for selection. The numeric percentage displayed at the top (e.g., 30\%) represents the proportion of the dataset included in the selected group. Players need to provide a group name at the top and click ‘Save’ to create a group or their preferences.}
  \Description{Game configuration: user preferences selection or group creation}
  \label{fig:Cgroup}
\end{figure}
% \vspace{-1em} % Reduce space after the figure

% \clearpage % Ensure new section starts on a new page

\subsection{Participants Demographics and Backgrounds}
\begin{figure}[htbp]
  \centering
  \begin{minipage}{0.5\textwidth}
    \centering
    \includegraphics[width=\linewidth]{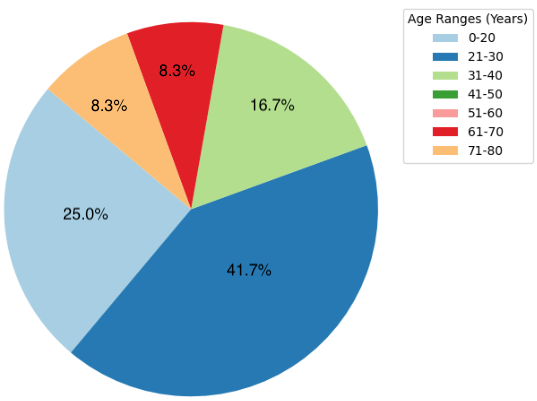}
    \caption{Participants' age range.}
    \label{fig:age}
  \end{minipage}%
  \hfill
  \begin{minipage}{0.45\textwidth}
    \centering
    \includegraphics[width=0.8\linewidth]{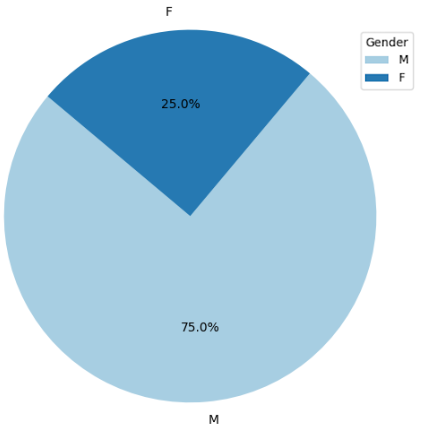}
    \caption{Participants' gender. Provided options included "Non-Binary" and "Prefer Not to Answer" as well.}
    \label{fig:gender}
  \end{minipage}
  
  \vspace{0.5cm} % Optional: Adds vertical space between rows of figures
  
  \begin{minipage}{0.5\textwidth}
    \centering
    \includegraphics[width=\linewidth]{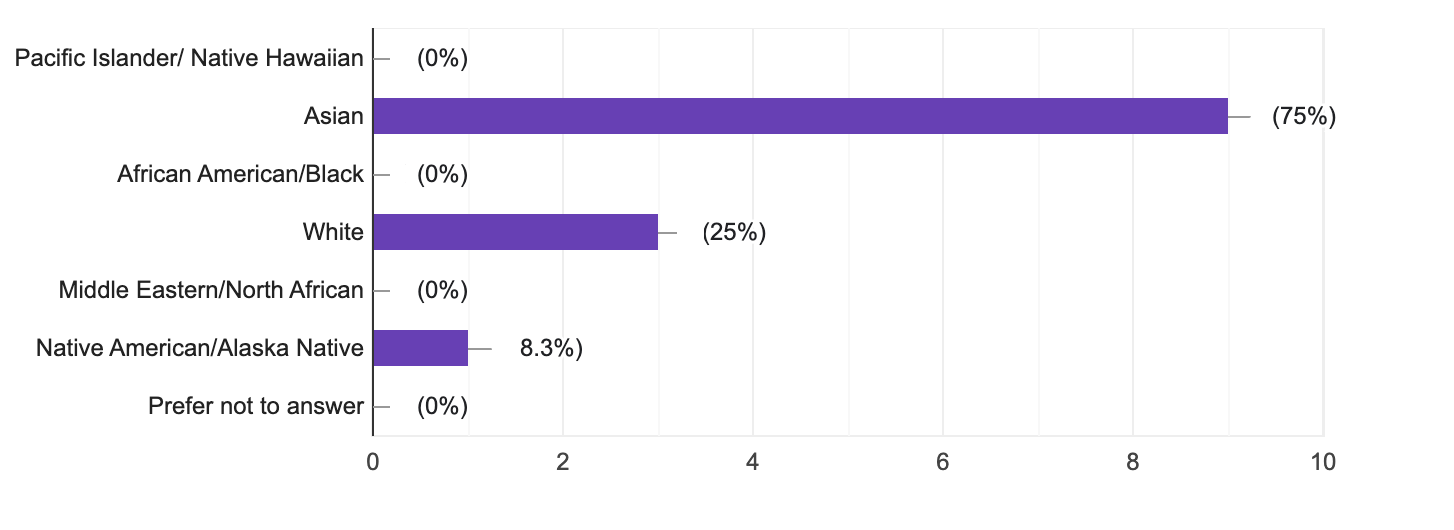}
    \caption{Participants' ethnicity.}
    \label{fig:ethnicity}
  \end{minipage}%
  \hfill
  \begin{minipage}{0.5\textwidth}
    \centering
    \includegraphics[width=\linewidth]{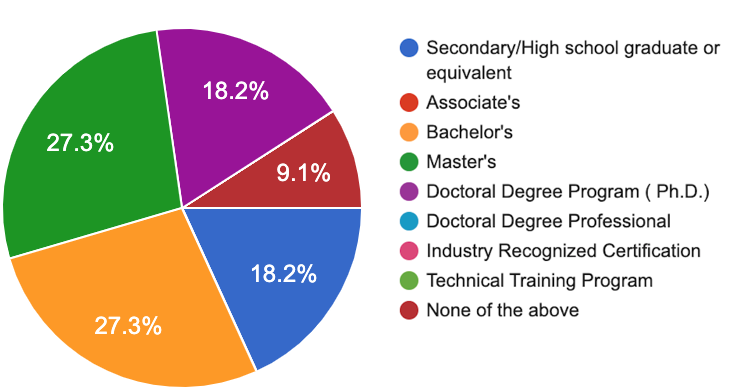}
    \caption{Participants' educational level.}
    \label{fig:education}
  \end{minipage}

  \vspace{0.5cm} % Optional: Adds vertical space between rows of figures
  
  \begin{minipage}{0.6\textwidth}
    \centering
    \includegraphics[width=\linewidth]{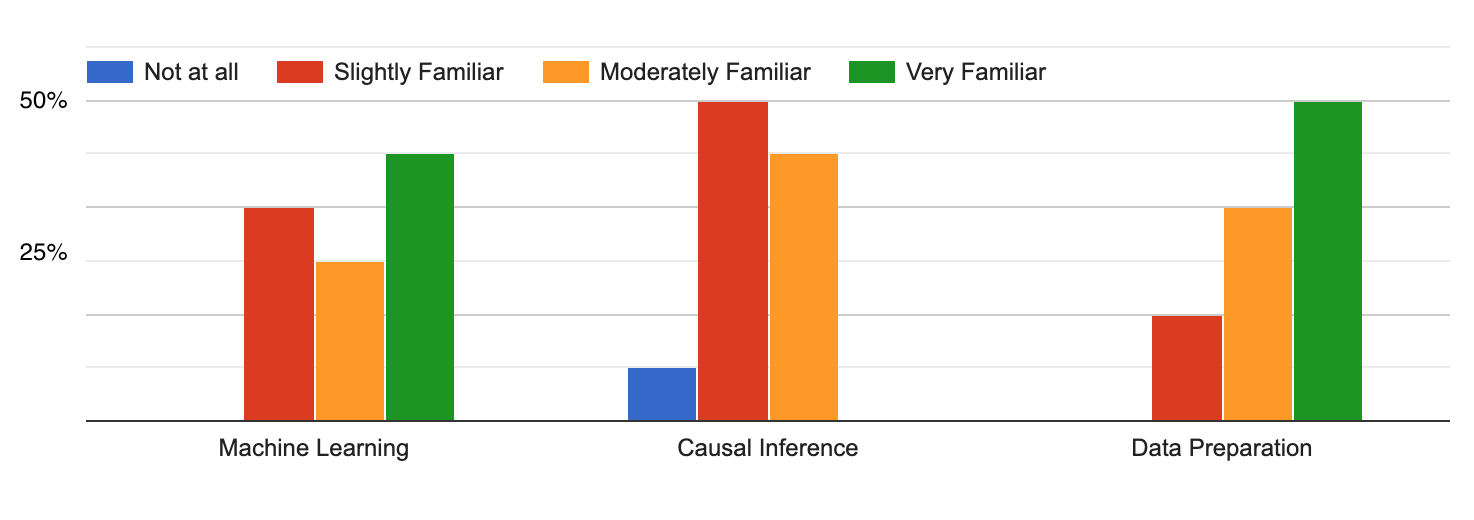}
    \caption{Participants' familiarity with related concepts.}
    \label{fig:familiarity}
  \end{minipage}
\end{figure}

\addcontentsline{toc}{subsection}{Participants Demographics and Backgrounds}
\label{appendix:users}
The demographic information of the users is provided in Figures \ref{fig:age},\ref{fig:gender} and \ref{fig:ethnicity}. Figures \ref{fig:education} and \ref{fig:familiarity} give some insights on participants backgrounds and familiarity with related concepts.

% \begin{figure}[H]
%   \centering
%   \includegraphics[width=0.7\linewidth]{figures/age_range.png}
%   \caption{Participants' age range.}
%   \label{fig:age}
% \end{figure}
% \begin{figure}[H]
%   \centering
%   \includegraphics[width=0.6\linewidth]{figures/gender.png}
%   \caption{Participants' gender. Provided options included "Non-Binary" and "Prefer Not to Answer" as well.}
%   \label{fig:gender}
% \end{figure}
% \begin{figure}[H]
%   \centering
%   \includegraphics[width=0.8\linewidth]{figures/nationality.png}
%   \caption{Participants' ethnicity.}
%   \label{fig:ethnicity}
% \end{figure}
% \begin{figure}[H]
%   \centering
%   \includegraphics[width=0.7\linewidth]{figures/education.png}
%   \caption{Participants' educational level.}
%   \label{fig:education}
% \end{figure}
% \begin{figure}[H]
%   \centering
%   \includegraphics[width=0.7\linewidth]{figures/familirity.png}
%   \caption{Participants' familiarity with related concepts.}
%   \label{fig:familiarity}
% \end{figure}

\subsection{Algorithm Computing the Game Metrics}
\addcontentsline{toc}{subsection}{Algorithm Computing the Game Metrics}
\label{appendix:algGM}

%% Algo 1
\begin{algorithm}[H]
\scriptsize
\caption{Compute Game Metrics}\label{alg:gm}
\begin{algorithmic}[1]

\State $group$ $\gets$ Object containing attributes, players care about.
\State $data$ $\gets$ DataFrame-like object containing tabular data including target variable.
\State $columns$ $\gets$ data.columns
% \State $jobStatus$ $\gets$ Aggregate job share for each attributes.

% \Require $group.length > 0$
% \Ensure $data.length > 0$

\Procedure{ComputingGroups}{}
    \State \textbf{Input:} $group$
    \State \textbf{Output:} Attribute wise percent care of each players
    \State \textbf{Initialize} attCare %$attCare \gets \{\}$
    
    \For{$player$ \textbf{from} $1$ \textbf{to} $group.length$} \Comment{1:n based index}
        % \State $attCare[player] \gets \{\}$
        \For{col, val \textbf{in} group[player].items()}
            \If{data[col] $==$ val}
                \State $attCare[key][val] \gets attCare[key][val] + 1$
            \EndIf
        \EndFor
    \EndFor
    % \State \Call{Definition2}{}  \Comment{Calling Definition2}
    \State \textbf{Return} $attCare$
\EndProcedure

\State \textbf{Initialize} groups
\State groups $\gets$ \Call{ComputingGroups}{}  \Comment{Calling ComputingGroups and assigning the result}

\Procedure{ComputingOutcome}{}
    \State \textbf{Input:} $data$
    \State \textbf{Output:} Attribute wise job distribution
    \State \textbf{Initialize} $attJob$
    
    \For{col \textbf{in} columns}
        \If{data[col] $==$ 0}
            \State $attJob[col][0] \gets data[data[col] == 0 \;\&\; data[data.target] == 1].shape[0]$
        \Else
            \State $attJob[col][1] \gets data[data[col] == 1 \;\&\; data[data.target] == 1].shape[0]$
        \EndIf
    \EndFor
    \State \textbf{Return} $attJob$
\EndProcedure

\State \textbf{Initialize} outcome
\State outcome $\gets$ \Call{ComputingOutcome}{}  \Comment{Calling ComputingOutcome and assigning the result}

\Procedure{ComputingAggregate}{}
    \State \textbf{Input:} $groups, outcome$
    \State \textbf{Output:} Attribute wise hiring diferences.
    \State \textbf{Initialize} $attAggregate $
    
    \For{$player$ \textbf{from} $1$ \textbf{to} $group.length$}
        \State $attAggregate[player] \gets groups[player] - outcome$
    \EndFor
    \State \textbf{Return} $attAggregate$
\EndProcedure

\State \textbf{Initialize} aggregate
\State aggregate $\gets$ \Call{ComputingAggregate}{}  \Comment{Calling ComputingAggregate and assigning the result}

\Procedure{ComputingTotaLossGain}{}
    \State \textbf{Input:} $groups, outcome$
    \State \textbf{Output:} Attribute wise hiring diferences.
    \State \textbf{Initialize} $attAggregate$
    
    \For{$player$ \textbf{from} $1$ \textbf{to} $group.length$}
        \State $attAggregate[player] \gets groups[player] - outcome$
    \EndFor
    \State \textbf{Return} $attAggregate$
\EndProcedure
\State \textbf{Initialize} totalLossGain
\State totalLossGain $\gets$ \Call{ComputingTotaLossGain}{}  \Comment{Calling ComputingTotaLossGain and assigning the result}

\State \textbf{Return} $groups, outcome, aggregate, totalLossGain$

\end{algorithmic}
\end{algorithm}

\clearpage % Ensure new section starts on a new page

\subsection{Player's Roles}
\begin{figure}[htbp]
    \centering
    \begin{subfigure}[b]{0.25\textwidth}
        \centering
        \includegraphics[width=\textwidth]{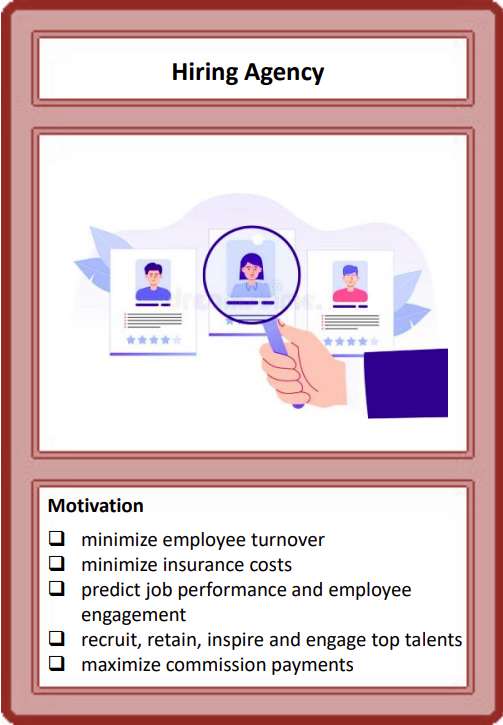}
        % \caption{Caption 1}
        % \label{fig:image1}
    \end{subfigure}
    \begin{subfigure}[b]{0.25\textwidth}
        \centering
        \includegraphics[width=\textwidth]{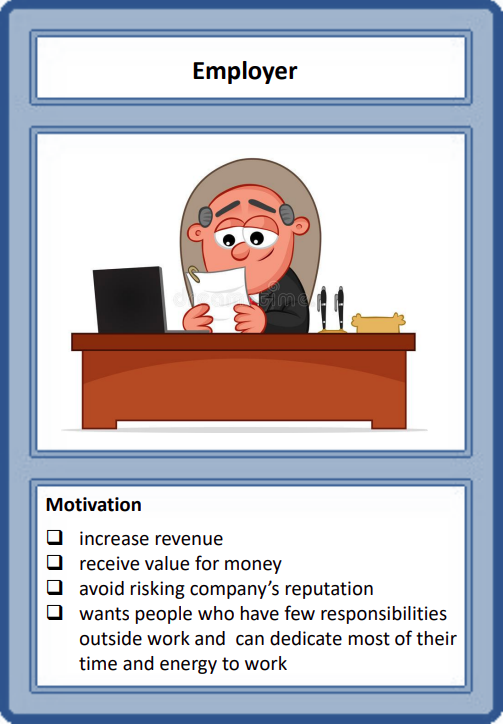}
        % \caption{Caption 2}
        % \label{fig:image2}
    \end{subfigure}
    \begin{subfigure}[b]{0.25\textwidth}
        \centering
        \includegraphics[width=\textwidth]{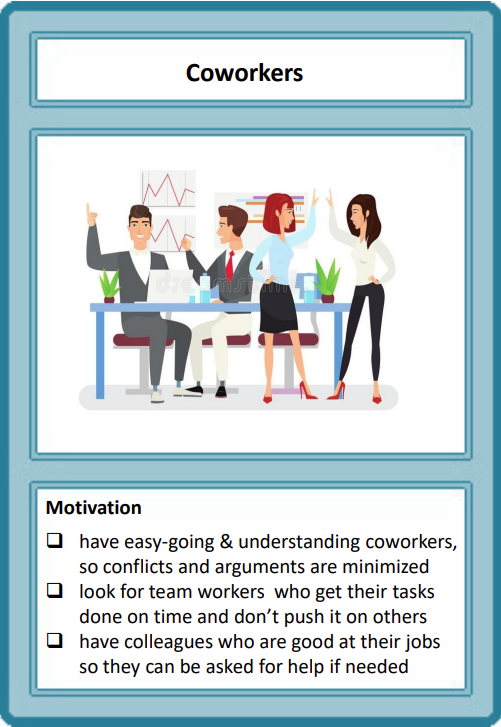}
        % \caption{Caption 3}
        % \label{fig:image3}
    \end{subfigure}
    
    \vspace{1em} % Add vertical space between rows
    
    \begin{subfigure}[b]{0.25\textwidth}
        \centering
        \includegraphics[width=\textwidth]{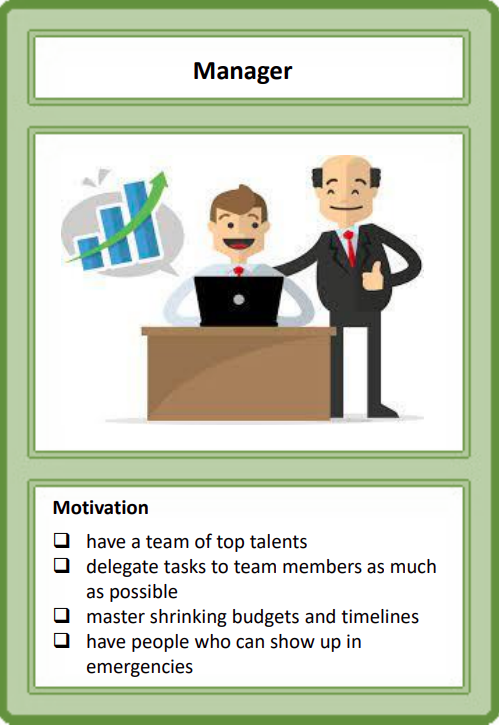}
        % \caption{Caption 4}
        % \label{fig:image4}
    \end{subfigure}
    \begin{subfigure}[b]{0.25\textwidth}
        \centering
        \includegraphics[width=\textwidth]{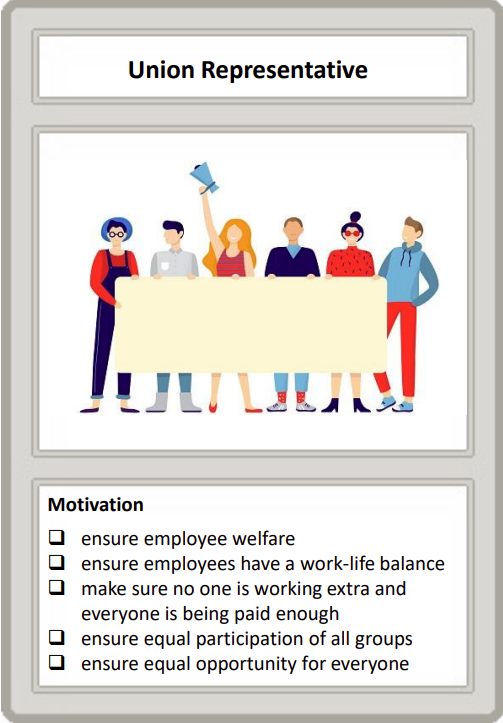}
        % \caption{Caption 5}
        % \label{fig:cards}
    \end{subfigure}
    \caption{Cards shown on the game interface to remind players of their goals for a particular role.}
    \label{fig:players cards}
\end{figure}

\addcontentsline{toc}{subsection}{Player's Roles}
The goals and objectives for each role were developed by analyzing job descriptions and requirements on recruitment websites such as LinkedIn and Indeed. The preferences selected for the players in our initial three user studies were aligned with the goals illustrated in Figure \ref{fig:players cards}. These preferences were not engineered to simplify reaching consensus by aligning the goals for all players. As indicated by the goals on the cards, some roles prioritize experience and talent, while others emphasize equal opportunities for all groups. There are goals and preferences that are aligned, as well as those that are in opposition, to ensure that the studies closely resemble real-world scenarios.
\label{appendix:cards}

\begin{table}[htbp]
\centering
\resizebox{\textwidth}{!}{
\small % Change the font size here
\begin{tabular}{|c|c|c|c|c|c|c|c|}
\hline
\textbf{Roles} & \textbf{Age} & \textbf{Gender} & \textbf{College Rank} & \textbf{Grade Point Average} & \textbf{Major} & \textbf{Work Experience} & \textbf{Race}\\ \hline
\textbf{Hiring Agency} & Both groups & \textemdash & Elite & Above 3 & Computer Science & Both groups & \textemdash \\ \hline
\textbf{Employer} & Above 42 & Both groups & \textemdash & Above 3 & \textemdash & Above 24 & Both Groups \\ \hline
\textbf{Manager} & Above 42 & Male & Elite & Above 3 & Computer Science & Above 24 & White \\ \hline
\textbf{Coworkers} & \textemdash & Both groups & Both groups & Above 3 & Both groups & Above 24 & \textemdash \\ \hline
\textbf{Union Rep.} & Both groups & Both groups & \textemdash & \textemdash & \textemdash & Both groups & Both Groups \\ \hline
\end{tabular}}
\caption{Preferences set for players in the first three user studies based on their roles. A line indicates that the role did not have any preference for that particular feature.}
\label{table:preferences}
\end{table}

\subsection{Analysis of Edge Strength Adjustments and Player Preferences}
\begin{figure}[H]
  \centering
  \includegraphics[width=0.75\linewidth]{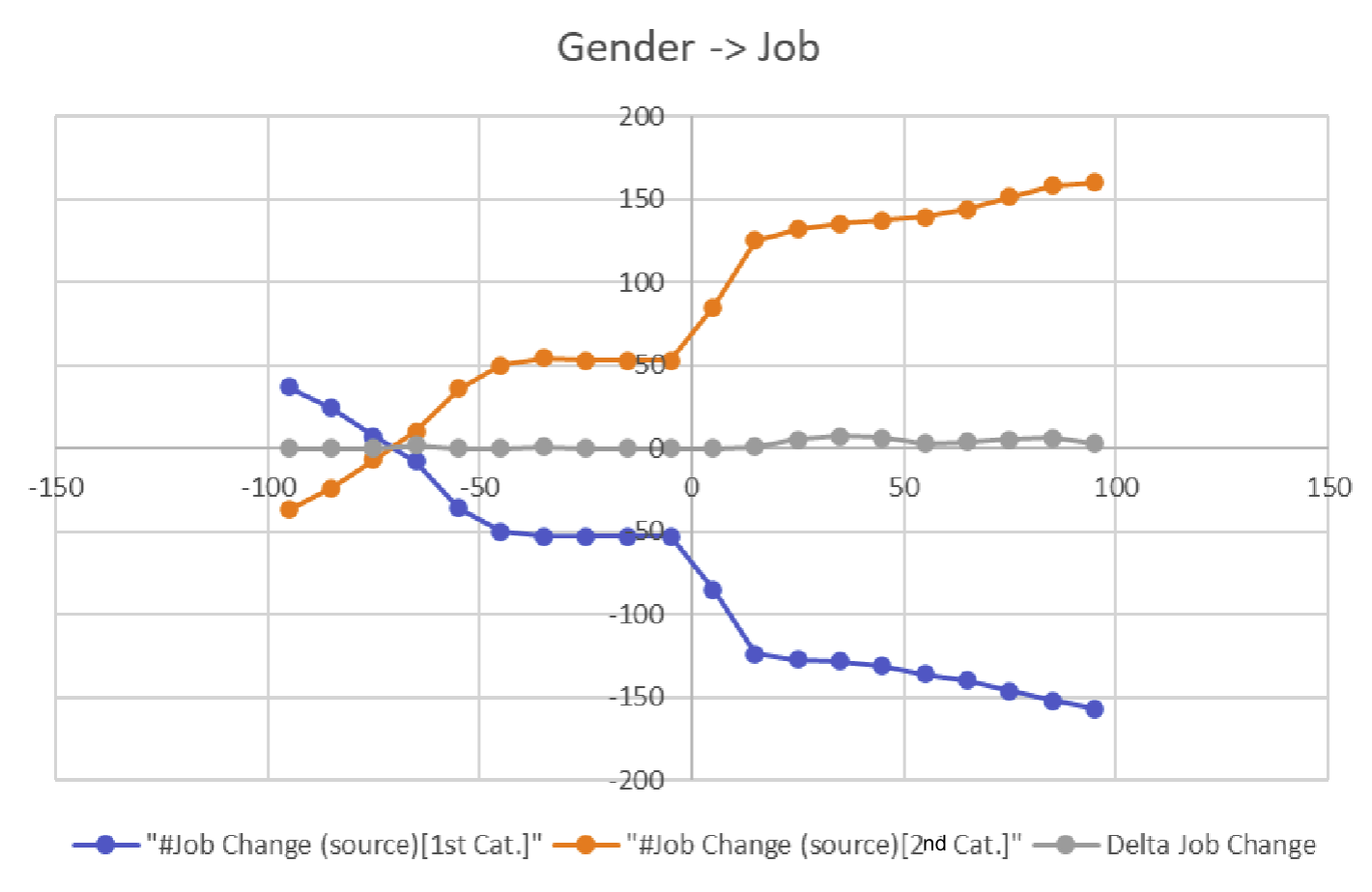}
  \caption{Impact of Edge Strength Adjustments on Job Distribution for the Gender -> Job Relationship. The x-axis represents the percentage change in edge strength, while the y-axis indicates the number of job changes. The blue line shows job changes for one gender category, the orange line for the other gender category, and the grey line represents the net difference in job changes.}
  \Description{Edge strength change and impact on job distribution}
  \label{fig:EPA}
\end{figure}

\addcontentsline{toc}{subsection}{Analysis of Edge Strength Adjustments and Player Preferences}
\label{appendix:AEPA}
% where -100\%  effectively removes the influence of that feature on the effect node, and +100\% doubles its strength.
The provided figure \ref{fig:EPA} illustrates the impact of adjusting the edge strength of the causal relationship between Gender and Job within the FairPlay framework. The x-axis represents the percentage change in edge strength, ranging from -100\% to +100\%. The y-axis indicates the number of job changes resulting from these adjustments. The blue and orange lines represent job changes for the two categories of gender, while the grey line shows the net difference in job changes, which remains around zero. This demonstrates that while jobs are redistributed between categories, the total number of jobs remains constant, preserving the system’s integrity. D-BIAS generates datasets by simulating data distributions via the causal network. As explained in \cite{ghai2022d}, when the weight of an edge in the causal diagram is decreased, it introduces more randomness into the corresponding antecedent node variable. For example, in figure \ref{fig:EPA}, decreasing the edge weight of Gender to Job (where Gender=1 corresponds to males and Job = 1 corresponds to getting the job) will result in a more balanced distribution of females getting the job, given all other qualities being equal.  Furthermore, by adding randomness instead of simply removing the sensitive Gender variable it also lowers proxy biases (if present) in variables downstream from them in the causal graph.

% The grey line remaining around zero demonstrates that while jobs are redistributed between genders, the total number of jobs remains constant, preserving the system’s integrity.

In the context of player preferences, adjustments favoring a particular category (e.g., Category 2) may lead to more jobs and game points for that category, negatively impacting players who prioritize the opposite category (e.g., Category 1). Thus, players must negotiate and find a balance that considers the goals of all participants. Changes in other variables within the causal network will also impact job distribution, requiring players to consider the broader network context. The point where the job change lines cross the x-axis represents the default job distribution for this variable, and this origin will shift based on the default distributions. This analysis assumes that other parts of the network remain static and are not altered during this specific analysis. This detailed analysis highlights how FairPlay enables users to explore the impact of edge strength adjustments on job distribution, maintaining coherence and balance in gameplay and aiding in informed decision-making for fairer job outcomes.

\subsection{The Concept of Preference Optimization in FairPlay \label{appendix:CPOF}}
\addcontentsline{toc}{subsection}{The Concept of Preference Optimization in FairPlay}

It is challenging to definitively determine whether fairness is objective or subjective. However, it is clear that fairness is often contested. This section explores potential approaches to address this issue.

\subsubsection{Concept Overview}

The principle underlying the FairPlay system is rooted in the notion of preference optimization, as introduced by Eliezer Yudkowsky \cite{yudkowsky2004coherent} in the context of AI safety. Unlike traditional approaches to reducing bias or enhancing fairness, FairPlay aligns the system's biases with the preferences of its users. This approach recognizes that what constitutes a "sensitive variable" or an appropriate decision criterion is ultimately a matter of user preference, rather than an objective standard.

\subsubsection{Analogy to Legislative Bodies}

The functioning of FairPlay can be likened to the dynamics within legislative bodies such as senates and parliaments. Representatives in these bodies do not follow a fixed, universally agreed-upon law-making algorithm. Instead, they evolve their strategies over time based on the preferences and interests of their constituents. This adaptive process mirrors how FairPlay aligns its outputs to user biases, allowing for shifts in priorities and strategies.

Just as legislators may oscillate between different principles and levels of engagement, FairPlay’s system is designed to adapt to the changing preferences of its users. The ultimate goal, similar to the preference for deliberation over dictatorship in legislative processes, is to ensure that the AI system reflects the collective preferences of its users rather than imposing a singular, potentially arbitrary standard of fairness.

\subsubsection{Implications for AI Alignment}

The concept of aligning AI systems to user preferences highlights a fundamental shift in how we think about fairness and bias in automated systems. It suggests that achieving fairness may be less about finding an objective measure of bias and more about ensuring that the system’s outputs are consistent with the values and preferences of those it serves. This approach acknowledges the complexity and variability of values and aims to create a more flexible and responsive AI system.

By situating the discussion of FairPlay within this broader context of preference optimization and legislative analogy, we underscore the importance of user-aligned AI and the potential limitations of traditional fairness metrics. This perspective not only informs the design and implementation of FairPlay but also contributes to the ongoing discourse on AI ethics and alignment.

%%%%%%%%%%%%%%%%%%%%%%%%%%%%%%%%%%
%%%%%%%%%%%%%%%%%%%%%%%%%%%%%%%%%%

\end{document}